\documentclass[smallcondensed]{svjour3}

\usepackage{amsmath}
\usepackage{graphicx}
\usepackage{subfig}
\usepackage{url}
\usepackage{hyperref}

\usepackage{amsmath,amssymb,amsfonts}
\usepackage{textcomp}
\usepackage{multirow}
\usepackage{multicol}
\usepackage{color,soul}
\setstcolor{red} 
\usepackage{natbib}

\usepackage{makecell}

\usepackage{pdflscape} 
\usepackage{pdfpages}
\usepackage{rotating}

\allowdisplaybreaks

\usepackage{caption}

\usepackage{booktabs}
\usepackage{tikz}
\usetikzlibrary{arrows,positioning}
\usepackage{lipsum, adjustbox}

\usepackage{setspace}

\newcommand{\STAB}[1]{\begin{tabular}{@{}c@{}}#1\end{tabular}}
    
\newcommand{\B}{\bfseries}

\begin{document}

\title{VARGAN: Variance Enforcing Network Enhanced GAN}
\titlerunning{VARGAN: Variance Enforcing Network Enhanced GAN}
\author{
        Sanaz Mohammadjafari\and
       Mucahit Cevik \and
        Ayse Basar
}

\institute{Sanaz Mohammadjafari \at
            Data Science Lab, Ryerson University, Toronto, Canada\\
             \email{sanaz.mohammadjafari@ryerson.ca}\\[-1.0em]
           \and
         Mucahit Cevik \at 
          Data Science Lab, Ryerson University, Toronto, Canada\\[-1.0em]
          \and
          Ayse Basar\at 
          Data Science Lab, Ryerson University, Toronto, Canada\\[-1.0em] 
}

\date{Received: date / Accepted: date}

\maketitle

\begin{abstract}
Generative adversarial networks (GANs) are one of the most widely used generative models. GANs can learn complex multi-modal distributions, and generate real-like samples. Despite the major success of GANs in generating synthetic data, they might suffer from unstable training process, and mode collapse.
In this paper, we introduce a new GAN architecture called variance enforcing GAN (VARGAN), which incorporates a third network to introduce diversity in the generated samples. 
The third network measures the diversity of the generated samples, which is used to penalize the generator's loss for low diversity samples. 
The network is trained on the available training data and undesired distributions with limited modality. 
On a set of synthetic and real-world image data, VARGAN generates a more diverse set of samples compared to the recent state-of-the-art models. 
High diversity and low computational complexity, as well as fast convergence, make VARGAN a promising model to alleviate mode collapse.


\end{abstract}
\keywords{Generative adversarial networks, mode collapse, multi-modal distribution, variance enforcing network}

\section{Introduction}

Generative Adversarial Networks (GANs) have emerged as a powerful generative model at learning data distributions. 
GANs have displayed a great potential in generating high quality images \citep{karras2017progressive,brock2018large}, and have been successfully used for image super-resolution \citep{bin2017high, ledig2017photo} and image-to-image translation \citep{isola2017image,zhu2017unpaired}. 
GANs consist of a generator network responsible for generating samples similar to the true distribution, and a discriminator network that discriminates between the samples from the true and generated distributions.
GANs aim to generate a set of samples that represent the true distribution, preserve the image quality and cover all the modes.
Achieving these goals rely on several factors such as model structure, network's objective functions, parameter tuning and training procedure. 
Although GANs have shown great potential in generating synthetic data, their unstable training procedure can cause non-convergence, and mode collapse \citep{salimans2016improved}.

Mode collapse refers to generating a limited number of modes from a multi-modal training data, and failing to generate a representative set of samples. 
This problem has limited GANs' potential in real-world applications such as MRI scan generation for brain segmentation \citep{mok2018learning} and electromagnetic engineered surface (EES) generation to improve the radio signal coverage in telecommunication \citep{mohammadjafari2021designing}. 
A large and growing body of literature has investigated various methods to alleviate the mode collapse issue in GANs through modified loss functions \citep{arjovsky2017wasserstein,che2016mode,gulrajani2017improved}, altering neural network structures \citep{zuo2019ldmgan,lin2018pacgan}, adapting different training methods \citep{DBLP:journals/corr/MetzPPS16} and latent space regularization \citep{gurumurthy2017deligan,li2021jdgan}. 
Regardless of the methodological differences, the proposed studies fall short of identifying a global solution.

In this work, we focus on the problem of mode collapse in GANs for image generation. Particularly, we are interested in binary and gray-scale images because of GANs' practical applications. 
For instance, effective usage of GANs in generating binary EES designs has been established in \citep{mohammadjafari2021designing}. However, cases of mode collapse were observed in the results, which affected the sample generation process. Moreover, research on image generation has been mostly restricted to colored image datasets. Therefore, our research can offer particular insights into binary and gray-scale image generation. Furthermore, binary and gray-scale image generation process requires a less complex model structure, and as a result, has lower computational complexity compared to the colored images. 
On the other hand, there could be certain challenges with binary images. 
For instance, category of binary EES designs changes rapidly by small changes in the image, which makes the training process of GANs for these binary images more challenging, requiring more research in this direction.



\paragraph{\textbf{Research goal}}
In this paper, we propose a novel GAN architecture called variance enforcing GAN (VARGAN) to alleviate the mode collapse problem in image generation, and increase the diversity of generated samples. 
We evaluate our method on two widely used datasets in the literature namely, 2D synthetic data and stacked MNIST, as well as on the real-life practical problem of generating EES designs.
Furthermore, considering the practical importance of conditional GANs, we examine the impact of conditioning on the mode collapse.

\paragraph{\textbf{Contributions}}
The main contribution of our paper is proposing a novel GAN architecture called VARGAN, which uses a third network called variance enforcing network (VarNet) to increase the number of generated modes, and reduce the mode collapse. Major contributions of our study can be summarized as follows:
\begin{itemize}
\item Our proposed method introduces a third network trained to compute the generated samples' diversity. VARGAN shows performance improvement over state-of-the-art GAN models in the literature. 

\item We explore the impact of conditioning and labels' length on mode collapse. 
We hypothesize that conditioning the GANs on auxiliary information can majorly impact the number of uniquely generated modes. 
The results show how VARGAN's performance, unlike other models, is not impacted by the length of conditioning labels.

\item We perform an extensive numerical study to evaluate GAN variants' performance on addressing the mode collapse using a variety of synthetic and real datasets. The results are analyzed based on various performance metrics employed in recent studies.
Specifically, we provide a detailed comparative analysis for different GAN architectures on a practical problem, EES generation, which has a significant application in telecommunication industry. Alleviating mode collapse problem and increasing the generated unique designs can greatly contribute to EES generation process by identifying designs that are not otherwise easy to obtain.
\end{itemize}

\paragraph{\textbf{Organization of the paper:}}
The rest of this paper is organized as follows. Section~\ref{sec:Background} summarizes the recent work on methods to address the mode collapse issue in GANs. In Section~\ref{sec:Methodology}, generic GAN structures are presented, and the methodology of the proposed architecture, as well as its training process are explained. Section~\ref{sec:Methodology} also covers the comparison of proposed method with the state-of-the-art GAN models. Section~\ref{sec:results} provides a discussion on the model performance for different types of datasets. Finally, the paper is concluded in Section~\ref{sec:conclusion}, and possible future directions are discussed.


\section{Background}\label{sec:Background}
In this section, we review recent studies on GANs with a special focus on mode collapse issue. A summary of the recent papers is reported in Table~\ref{tbl:lit_rev}, which contains information on the methodology, GAN structures and datasets used in the numerical experiments.

\citet{salimans2016improved} proposed several approaches based on modifications in GANs' architecture and loss functions. 
Feature matching, historical averaging and mini-batch discrimination are examples of the methods proposed to improve the convergence and mode collapse in GANs. 
Mini-batch discrimination approach adds a new component to the last layer of discriminator by computing the distance among the last layer's samples. 
The authors claim that closer samples are more likely to be created by the generator, and their proposed enhancements help with discriminator's decision making and improves GAN's convergence.


\clearpage
\newpage
\thispagestyle{plain}
\begin{landscape}
\setlength{\tabcolsep}{8pt}
\renewcommand*{\arraystretch}{1.35}
\begin{table}[!ht]
\begin{center}
\caption{Summary of the relevant papers addressing the mode collapse
{\footnotesize {(FF: Feed-forward, ConvGAN: Convolutional GAN, cCGAN: Conditional Convolutional GAN)}}.}
\label{tbl:lit_rev}
\scalebox{0.77}{

\begin{tabular}{lllll}
\toprule
Paper                 & Proposed method  & Methodology & GAN structure & Dataset  \\ \midrule
\multirow{3}{*}{\citet{salimans2016improved}} & \multirow{3}{*}{Minibatch discrimination} &  \multirow{3}{*}{\begin{tabular}{@{}l@{}l@{}}Distance of extracted features \\ in the last layer of discriminator \\ added to discriminator's loss as penalty \end{tabular}}  & \multirow{3}{*}{ConvGAN} & MNIST$^\dagger$\\
& & & & CIFAR-10\\
& & & & \\
\hline
\multirow{2}{*}{\citet{che2016mode}} & \multirow{2}{*}{Mode regularized GAN} & \multirow{2}{*}{\begin{tabular}{@{}l@{}}Jointly train an encoder\\ add penalty factors to generator's loss function\end{tabular}} & \multirow{2}{*}{ConvGAN} & Stacked MNIST\\
& & & & CelebA\\
\hline
\multicolumn{1}{l}{\citet{arjovsky2017wasserstein}} & Wasserstein GAN & Wasserstein distance plus weight clipping & ConvGAN & LSUN-Bedrooms\\
\hline
\multirow{2}{*}{\citet{gulrajani2017improved}} &\multirow{2}{*}{WGAN-GP} & \multirow{2}{*}{\begin{tabular}{@{}l@{}}Replace weight clipping in Wasserstein GANs\\ with an enforced Lipschitz constraint\end{tabular}}& \multirow{2}{*}{ConvGAN} & LSUN-Bedrooms\\
& & & & CIFAR-10\\
\hline
 
\multirow{3}{*}{\citet{tolstikhin2017adagan}} & \multirow{3}{*}{AdaGAN} & \multirow{3}{*}{Mixture of GANs trained on reweighted samples} & ConvGAN & MNIST$^\dagger$ \\
& & & & Stacked MNIST\\
& & & FF GAN & Synthetic data$^\dagger$ \\
\hline
\multirow{2}{*}{\citet{park2018megan}} &\multirow{2}{*}{MEGAN} & \multirow{2}{*}{Mixture of GANs} & \multirow{2}{*}{ConvGAN} & CelebA\\
&&&& LSUN-Church outdoor\\
\hline

\multirow{3}{*}{\citet{lin2018pacgan}} & \multirow{3}{*}{PacGAN} & \multirow{3}{*}{Discriminator's architecture receives $m$ samples at a time} & ConvGAN & Stacked MNIST\\
& & & & CelebA\\
& & & FF GAN & Synthetic data$^\dagger$ \\
\hline
\multirow{5}{*}{\citet{ghosh2018multi}} & \multirow{5}{*}{MADGAN} & \multirow{5}{*}{Multi-generators} & FF GAN & Synthetic data\\
&&& ConvGAN & Stacked MNIST\\
&&&  & CelebA\\
&&& cCGAN & Night-to-day\\
&&&& Edges-to-handbags\\

\hline

\multirow{3}{*}{\citet{elfeki2019gdpp}} & \multirow{3}{*}{GDPP} & \multirow{3}{*}{\begin{tabular}{@{}l@{}}Use negative correlations within a subset\\ as diversity measure added to generator's loss\end{tabular}} & FF GAN & Synthetic data\\
&&& ConvGAN & Stacked MNIST\\
&&&& CIFAR-10\\
\hline

\multirow{3}{*}{\citet{mao2019mode}} & \multirow{3}{*}{MSGAN} &
\multirow{3}{*}{\begin{tabular}{@{}l@{}}Image difference divided by noise difference\\ added to generator's loss function\end{tabular}}& Conditioned on labels & CIFAR-10\\
&&& Conditioned on Images & Winter-to-summer\\
&&& Text to image synthesis & CUB-200-2011\\
\hline
\multirow{3}{*}{Our study} & \multirow{3}{*}{VARGAN} &
\multirow{3}{*}{\begin{tabular}{@{}l@{}} A third network trained on samples' diversity\\ adds a penalty to the generator's loss to encourage diversity \end{tabular}}& FF GAN& Synthetic data$^\dagger$\\
&&& FF GAN, ConvGAN & Stacked MNIST\\
&&&& EES$^{\textsubscript{*} \dagger}$\\
\bottomrule
\multicolumn{2}{l}{$^{\textsubscript{*}}$ propriety dataset}\\
\multicolumn{2}{l}{$^\dagger$ gray scale images}\\
\end{tabular}
}
\end{center}
\end{table}
\end{landscape}

Altering the loss function to stabilize GANs' training, and to increase the covered modes is investigated in various studies. These methods include introducing new distance metrics such as Wasserstein distance \citep{arjovsky2017wasserstein}, incorporating regularization penalties for cost functions of the networks \citep{che2016mode,gulrajani2017improved}, and encouraging diversity in generated samples through an unsupervised penalty loss based on the samples in last layer of the  discriminator \citep{elfeki2019gdpp}. 
Previous studies often evaluated the impact of modifications in neural networks' structures by changing GANs' networks architectures, and incorporation of multiple networks to cover the missed modes  \citep{park2018megan,hoang2018mgan,srivastava2017veegan,zhong2019rethinking}. For instance, \citet{lin2018pacgan} altered the discriminator's architecture to receive $m$ merged samples of the data with the same label (fake or real) and generate one label instead of $m$. Their proposed method, PacGAN, showed great improvement in fake samples' diversity.
GANs construction using multiple networks is inspired by the theory established for mode collapse in \citep{arjovsky2017towards} where disjoint distribution of real and generated data is considered as the source of instability and missed modes. Multi-agent GAN (MADGAN) is one particular approach that contains multiple generators and one discriminator network to encourage different generators toward separate modes of the data, and increase the variety of generated samples  \citep{ghosh2018multi}. \citet{tolstikhin2017adagan} proposed adaptive GAN (AdaGAN), which incrementally adds a new model to a mixture of GANs, and evaluates a new GAN model on re-weighted samples.

Majority of the studies investigating mode collapse consider vanilla GANs structures and exclude conditional GANs as shown in Table~\ref{tbl:lit_rev}. However, conditional GAN architectures have a wide range of applications in many areas \citep{isola2017image, zhu2017unpaired, mohammadjafari2021designing}. They also suffer from mode collapse, which has been investigated in a limited number of studies. \citet{zhu2017multimodal} proposed a new hybrid GAN structure namely, Bicycle GAN, which incorporates bidirectional mapping from latent code to output. This study was implemented on particular conditional tasks, and has a significant computational complexity overhead. 
\citet{mao2019mode} proposed a new general approach called MSGAN by adding a regularization term to the generator's loss function that is applicable to various GAN architectures. 
MSGAN approach encourages the generator to map two different samples conditioned on the same context to be as distant from each other as possible.
 
Our VARGAN method utilizes some of the aforementioned enhancements such as modified loss functions and adding a new penalty term to the generator's loss (e.g., see \citep{elfeki2019gdpp}), which computes the generated samples' diversity. 
However, in contrast to the previous approaches, our proposed penalty term is calculated by a third network trained on various sets of training samples with their relative diversity level. 
Different than multi-network approaches, our method does not incorporate a new generator or a discriminator, and the third network is solely responsible to provide feedback to the generator.  
Moreover, our approach incorporates the same strategy of modified structures as \citep{lin2018pacgan} for its new network's architecture. 

\section{Methodology}\label{sec:Methodology}
In this section, we describe our methods, datasets and the experimental setup. We first review the structure of the vanilla GAN and its training procedure, and then explain the VARGAN architecture in detail.
In addition, we discuss how our proposed framework compares to the existing GAN architectures.

\subsection{Standard GAN models}
\begin{figure}[!ht]
\centering
\includegraphics[width=0.8\textwidth]{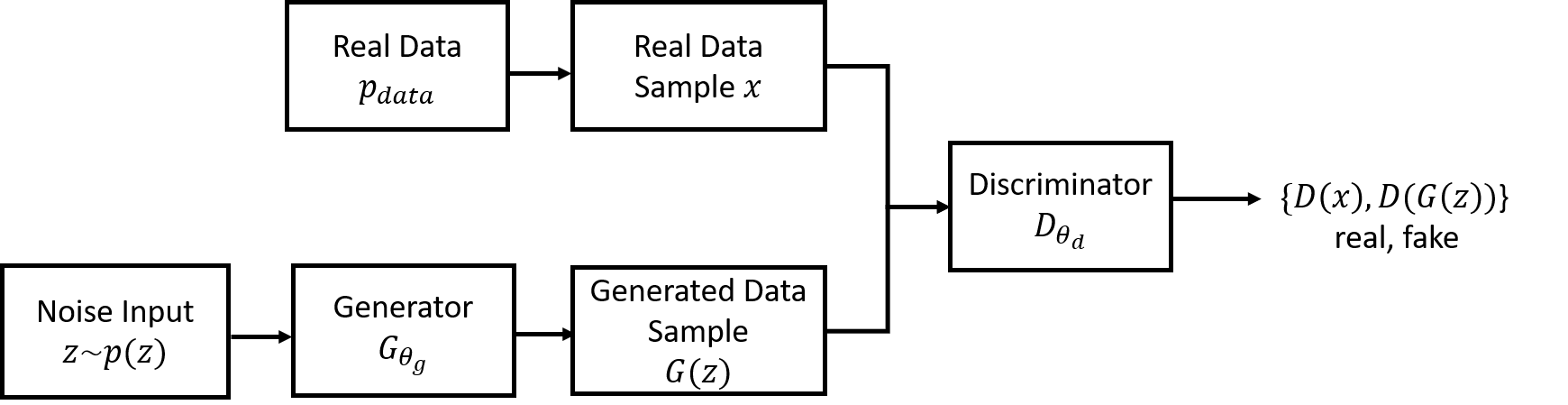}
\caption{GAN structure consists of two competing networks, namely, discriminator and generator}
\label{fig:GAN_structure}
\end{figure}

A generic GAN architecture consists of two networks known as the generator and the discriminator, both competing against each other (see Figure~\ref{fig:GAN_structure}). The generator learns the distribution $p_g$ from a uniform or normal distribution input noise $z$ mapped through the function $G(z; \theta_g)$ to the samples. $G(z; \theta_g)$ is a differentiable function, usually defined as a neural network with a set of parameters $\theta_g$. The discriminator $D(x,\theta_d)$, which is also a differentiable function with parameters $\theta_d$, maps input samples to a probability value representing whether each sample is real or generated. Both networks are trained simultaneously, where discriminator aims to maximize and generator tries to minimize the objective value. Therefore, GAN is modelled as a min-max game with a value function $V(D,G)$, that is
\begin{equation}
     \label{eq:GANS-formula}
    \min_G{\max_D{V(D,G)}} = E_{x\sim p_{data}(x)}[\log D(x)] + E_{z\sim p_{z}(z)}[\log(1-D(G(z)))]
\end{equation}


\begin{figure}[!ht]
\centering
\subfloat[Discriminator training \label{fig:GAN_disc_training}]{\includegraphics[width=0.8\textwidth]{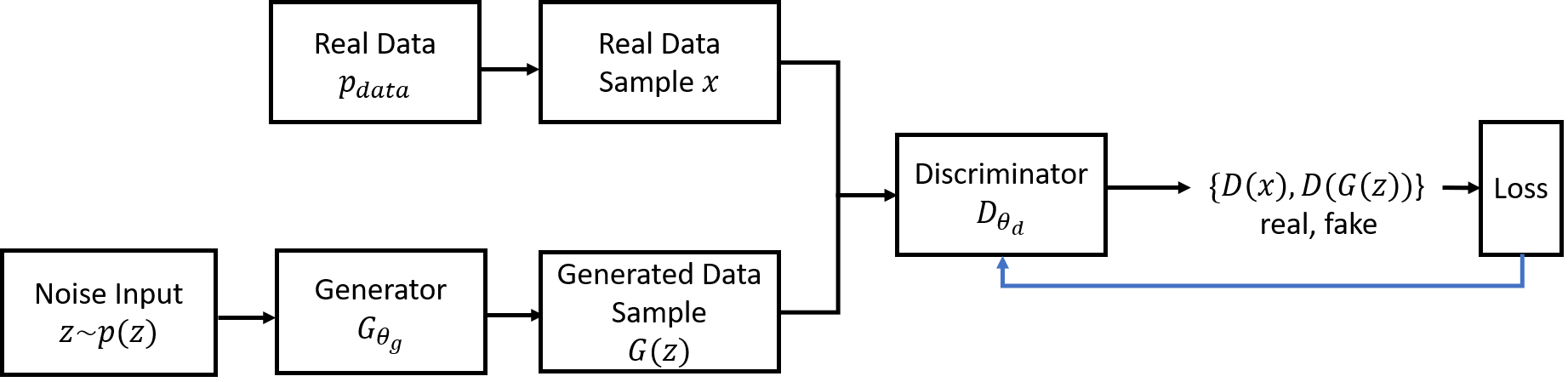}}
\hfill
\subfloat[Generator training \label{fig:GAN_gen_training}]{\includegraphics[width=0.8\textwidth]{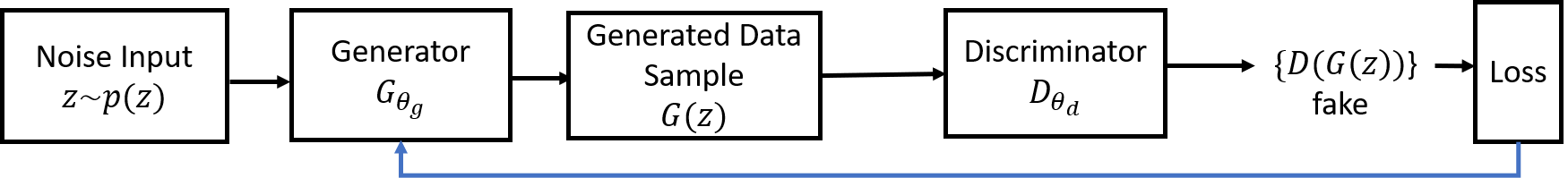}}
\caption{GAN model training procedure including discriminator and generator feedback signals}
\label{fig:GAN_model_training}
\end{figure}

The discriminator aims to maximize the value function by generating a probability value of one for the samples from the real data $p_{data}$ and zero for the samples from the generated distribution $p_g$. 
On the other hand, generator minimizes the objective value by trying to trick the discriminator into outputting one for the generated samples. 
The training procedure of the GANs' networks is illustrated in Figure~\ref{fig:GAN_model_training}.

Vanilla GAN has no control on the category of the generated samples. Therefore, conditional GANs were proposed to incorporate supplemental data as labels in both the generator and discriminator, and generate samples of a desirable category \citep{mirza2014conditional}. Convolutional GANs are another important variant of GANs that use convolutional neural network (CNN) instead of feed-forward structure in their networks \citep{radford2015unsupervised}. 

\subsection{VARGAN}
We propose a new GAN framework called VARGAN illustrated in Figure~\ref{fig:VARGAN_structure} by introducing a third network called VarNet to the vanilla GAN structure. 
The VarNet is used to reduce the mode collapse by encouraging diversity in the generated samples, which results in increased number of generated modes. Below, we describe VARGAN architecture and the training procedure.
\begin{figure}[!ht]
\centering
\label{fig:VARGAN_structure}
{\includegraphics[width=0.8\textwidth]{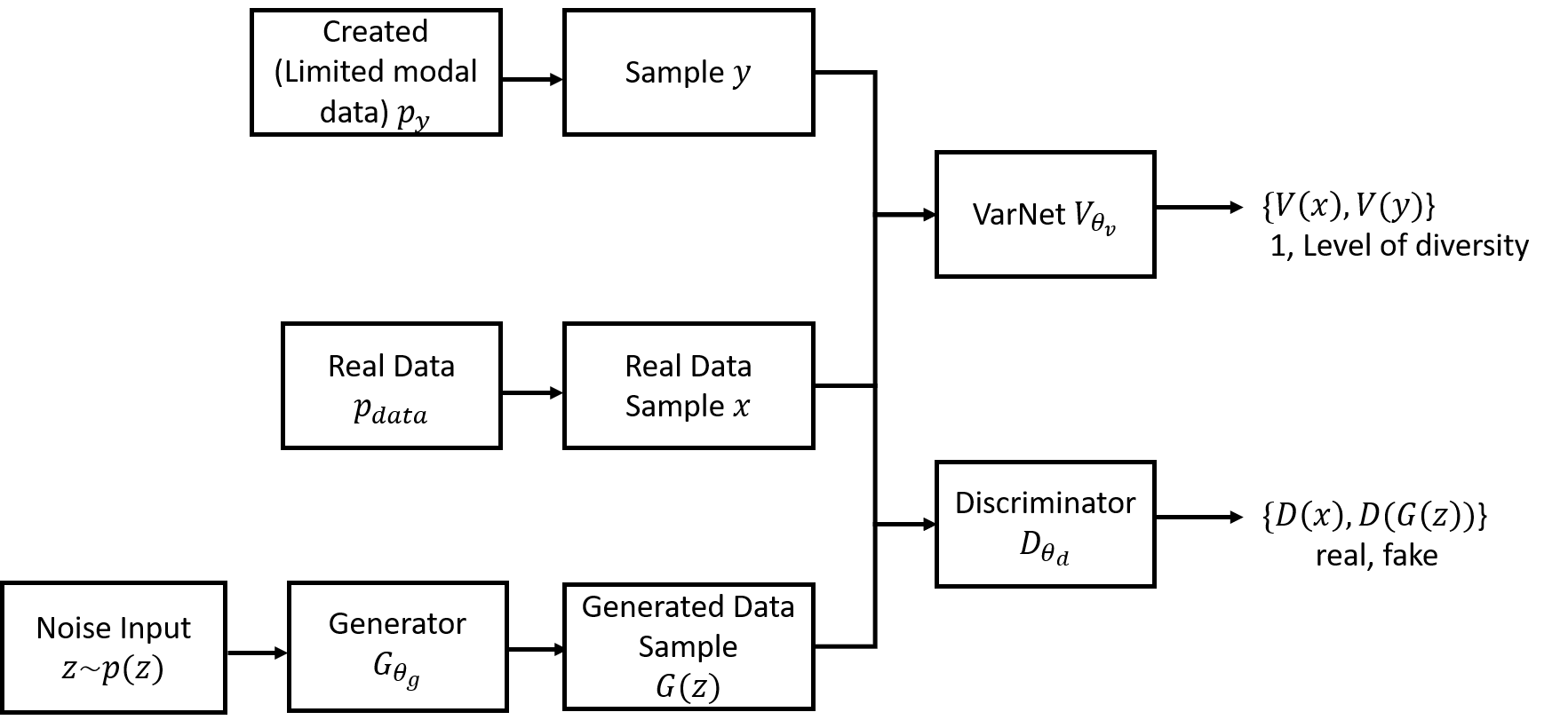}}
\caption{VARGAN structure including generator, discriminator and VarNet networks}
\label{fig:VARGAN_structure}
\end{figure}

\subsubsection{Data samples}
Discriminator network in GANs utilizes two sets of data for its training process, namely, the real and fake data. 
GAN aims to learn the real data distribution $p_{data}$ with $N$ number of unique modes. 
Therefore, $N$ is the target and highest number of unique modes in $p_{data}$ that a generator can create. 
It is important to note that for some datasets with unknown number of unique modes, we can select different $N$ values to encourage the diversity. 
Fake data distribution $p_g$ is created by the generator network that needs to improve in quality and diversity to reach the true distribution.

In the VARGAN framework, we introduce another set of data called limited modality data $p_y$, which defines the undesirable distributions with restricted number of modes. 
The limited modality data is used for VarNet training process. 
The undesirable distributions include lower number of modes than the target data with different sample distributions. 
Different than real and fake data, limited modality data is created by following a specific set of rules and using the real data.

The first step in creating $p_y$ is to define the number of modes $n<N$ that data sample covers, which is less than the target number of modes. Then, we select $n$ different data points from the batch of training data and repeat each of them $\frac{B}{n}$ times to create a uniform set of $n$ different categories in the batch, where $B$ represents the batch size. 
Each created batch of data with $n$ different modes has a relative Mode Coverage Ratio (MCR) value between zero and one, which is defined as follows:
\begin{equation}
    \label{eq:mcr-formula}
      \text{MCR} =  
      \begin{cases}
      \frac{\mathrm{L}}{\displaystyle \mathrm{1} +  S_1 \cdot e^{-\frac{n}{N}  \cdot S_2}}  & n = 2,\hdots, N-1, \enspace L \in (0,1], \enspace n \mid B\\ 
      
      0 & n = 1\\
      
    \end{cases}  
\end{equation}

\begin{figure}[!ht]
\centering
\subfloat[MCR formulation trajectory with $L=1$ \label{fig:MCR_trajectory_L_1}]{\includegraphics[width=0.8\textwidth]{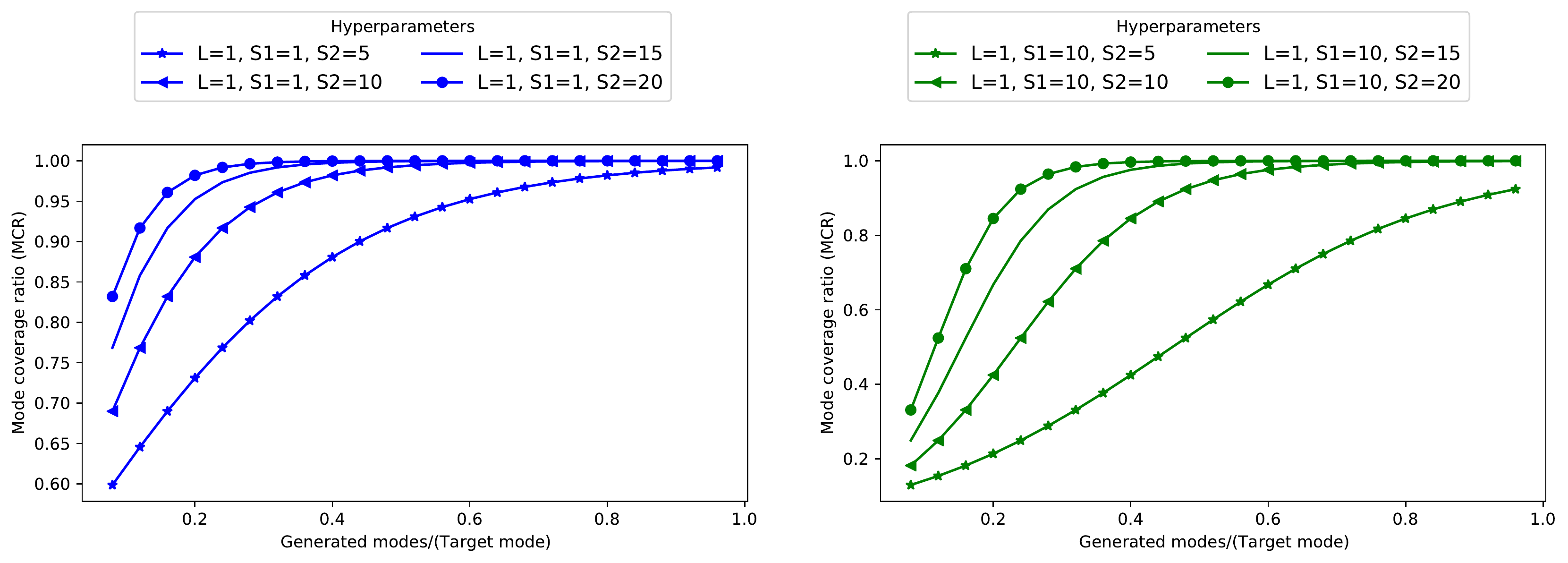}}
\hfill
\subfloat[MCR formulation trajectory with $L=0.5$ \label{fig:MCR_trajectory_L_05}]{\includegraphics[width=0.8\textwidth]{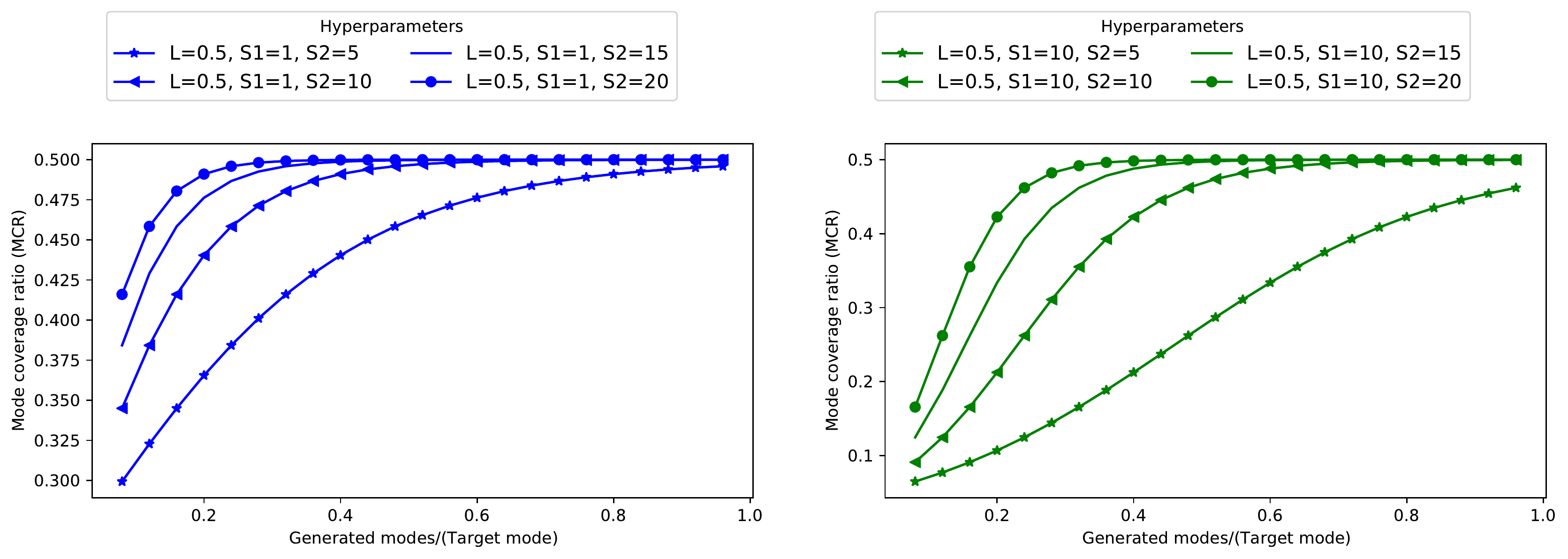}}
\caption{MCR formulation trajectory with respect to the number of generated modes divided by the target number of modes using different hyperparameter values for $L$, $S_1$ and $S_2$}
\label{fig:MCR_trajectory}
\end{figure}

MCR value of zero indicates a single unique design in the sample, representing the lowest level of diversity. As the number of modes increase, the MCR value increases as well. 
Equation~\eqref{eq:mcr-formula} shows the MCR value as a function of the modes ($n$) divided by the target number of modes ($N$), representing the percentage of generated modes. Parameters $L$, $S_1$ and $S_2$ are the constants that control the MCR trajectory. 
Comparison of Figure~\ref{fig:MCR_trajectory_L_05} and~\ref{fig:MCR_trajectory_L_1} shows how $L$ controls the final value of MCR for high number of modes by using two different values for $L$.
Figure~\ref{fig:MCR_trajectory} also illustrates how $S_1$ controls the initial value of MCR. 
A small MCR value for low number of covered modes can increase the generator's loss at the beginning steps of the training process. 
This can adversely affect the model convergence, and needs to be controlled. 
Moreover, a high MCR value for low number of modes can prevent the generator from further increasing the diversity. 
Parameter $S_2$ controls the slope of the MCR, and since slope affects the model's convergence process, $S_2$ affects the convergence to the target MCR value.
These constants are determined through preliminary analysis (see details in Appendix~\ref{ap:Limited_data_formula}). 
Generation of limited modality data and the relative MCR remains a challenging part of the VARGAN framework, since it enforces a few constraints on the number of selected modes. 
One constraint is that the number of modes should divide the batch size because of the VarNet structure explained in the following sections. 
 
\subsubsection{VARGAN architecture}
\begin{figure}[!ht]
    \centering
    \includegraphics[width=0.4\textwidth]{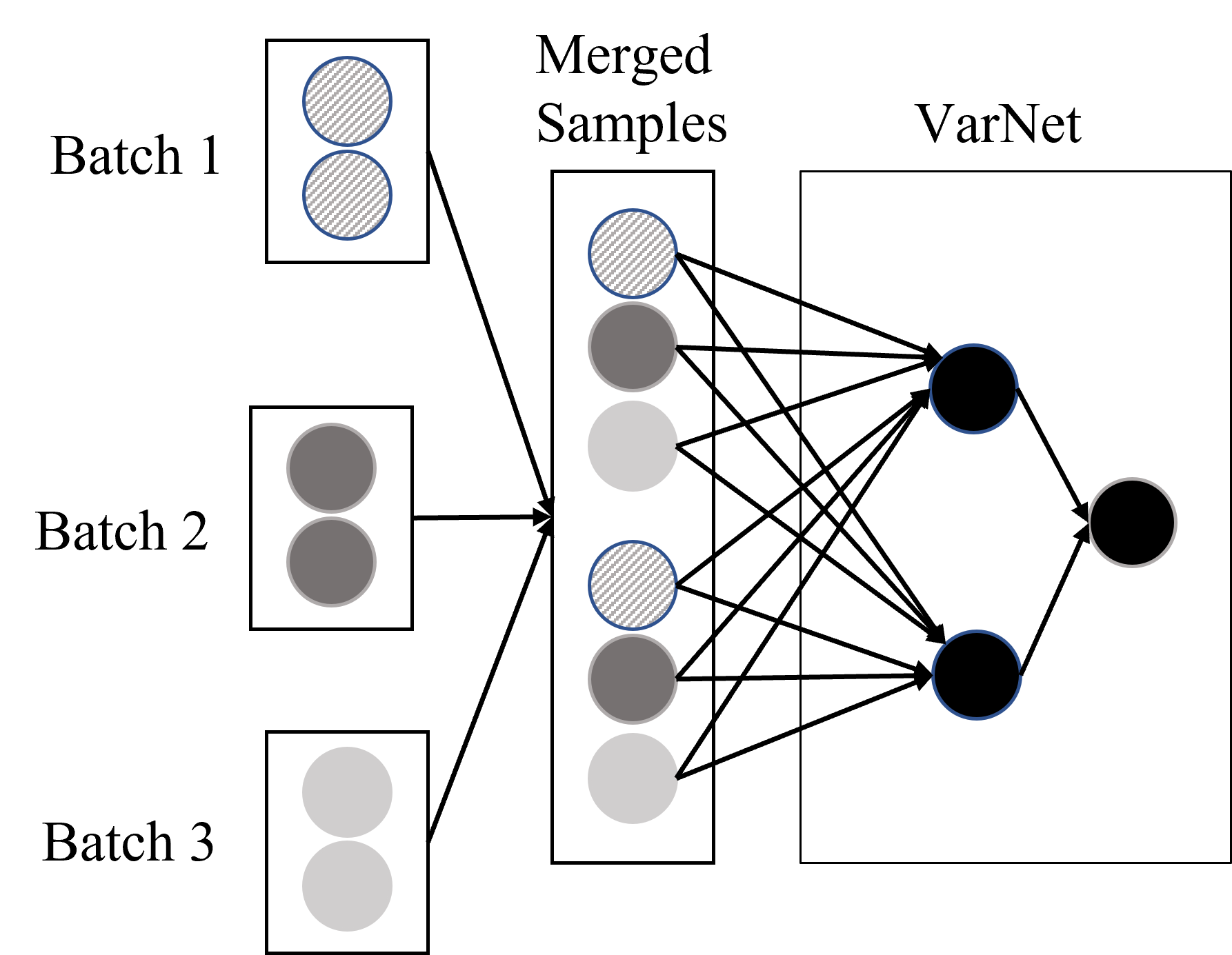}
    \caption{Limited modality data with $n=3$ and $B=6$ passed to a sample VarNet structure with $m=6$ with one output}
    \label{fig:limited_data}
\end{figure}

The VARGAN framework maintains the same discriminator and generator architecture as the standard GANs. 
The new network, VarNet $V(x,\theta_v)$ is a differentiable function created by a neural network with $\theta_v$ parameters that associates each set of input samples to an MCR value. 
As shown in Figure~\ref{fig:VARGAN_structure}, the VarNet receives a sample data $x$ from real data distribution $p_{data}$, and a sample data $y$ from limited modality data $p_y$, and maps them to a single scalar value between zero and one representing the level of diversity of the samples. 
The VarNet network employs the PacGAN structure \citep{lin2018pacgan} with a degree of packing equal to the batch size. 
The packing degree $m$ refers to the number of augmented samples with the same label. 
In our case, packing degree is equal to the batch number indicating that VarNet  receives the whole batch of the data $V(X_1, X_2, \hdots, X_m)$ and maps it to one MCR value. 
Note that since packing number $m$ is equal to $B$, $n$ needs to divide batch size $B$ to have a reasonable MCR value for each batch. 
A sample of limited data set with $n=3$ and batch size of six passed to a VarNet is illustrated in Figure~\ref{fig:limited_data}.

\subsubsection{VARGAN training process}
\begin{figure}[!ht]
\centering
\subfloat[VarNet training \label{fig:VarNet_training}]{\includegraphics[width=0.7\textwidth]{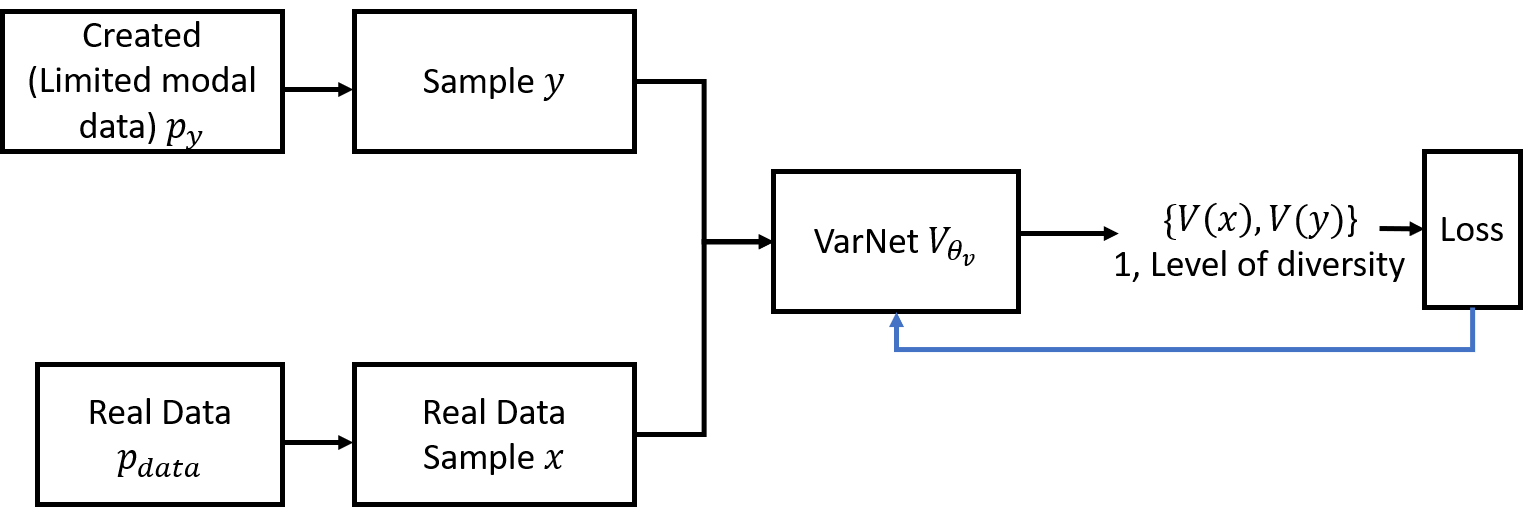}}

\subfloat[Generator training \label{fig:gen_training}]{\includegraphics[width=0.8\textwidth]{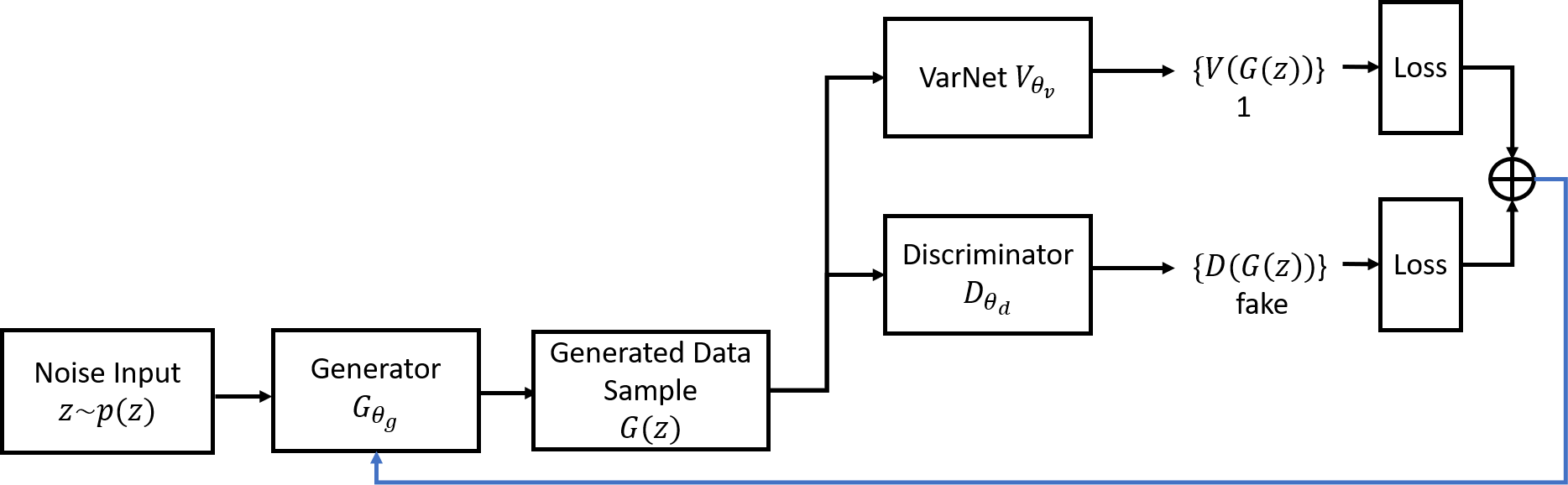}}
\caption{Training procedures VARGAN networks including VarNet and generator (discriminator's training process is similar to vanilla GAN and is not illustrated). }
\label{fig:vargan_model_training}
\end{figure}

The VARGAN training procedure is presented in Figure~\ref{fig:vargan_model_training}.
The loss function for VarNet training is as follows:
\begin{equation}
    \label{eq:V-formula}
    \min_V{F(V)} = E_{x\sim p_{data}(x)}[\log(1- V(x))] 
    + E_{y\sim p_{y}(y)}[\log(MCR-V(y))]
\end{equation}
The network receives two sets of inputs for the training process. The first one is a sample of training data $p(x)$ with the target distribution and a high MCR value as labels. The second input is a set of limited modality data $p(y)$ with different MCR values. 
The VarNet aims to minimize loss value by mapping the training data samples to label one and the limited modality data to the relative MCR. 
The structure and loss function of the discriminator for VARGAN remain the same compared to generic GAN architecture (see Figure~\ref{fig:GAN_disc_training}). 
However, the generator's loss function has a new addition provided in Equation~\eqref{eq:G-formula} as follows:
\begin{align}
    \label{eq:G-formula}
    \min_G{F(D,G,V)} &= E_{z\sim p_{z}(z)}[\log(1-D(G(z)))] \\
     &+ E_{z\sim p_{z}(z)}[\log(1-V(G(z)))] \cdot C
     \nonumber
\end{align}
The new term calculates the loss between the target level of MCR, which is one, and the MCR of the generator's outputs measured by VarNet. 
This term is also multiplied with a positive coefficient factor $C \leq 1$, which controls the impact of VarNet on generator's loss. 
The value of parameter $C$ is selected based on hyperparameter tuning experiments and for the majority of the experiments, it is found to be one.

\subsubsection{Comparison with other architectures}
VARGAN employs similar strategies to other available GAN architectures to deal with mode collapse issue. 
Our proposed framework incorporates the generated samples diversity in each iteration to penalize the generator towards a multi-modal output.
The idea of penalizing the generator for the diversity is also discussed in \citep{elfeki2019gdpp} where Determinantal Point Process (DPP) is used to compute the diversity from the features of the discriminator's last layer. 
However, in our work, a third network is trained to evaluate the samples diversity. 
Also note that our approach differs from multi-network models. 
VARGAN does not rely on any additional generator or discriminator networks. 
The third network (VarNet) sends an additional feedback to the generator to help with the diversity.
We incorporate the same methodology of modified structures as \citep{lin2018pacgan} for the VarNet architecture to help the model differentiate between samples. 
However, the main contribution of our model to address the mode collapse issue is the additional penalty term added to the generator. 
Moreover, mini-batch discrimination \citep{salimans2016improved} incorporates a distance measure in the discriminator architecture to help distinguish between real and fake samples, which is not employed in our VARGAN architecture. 

\subsection{Datasets}
We consider three particular datasets in our numerical study, which are described below.
\begin{itemize}

   \item \emph{Synthetic data:} We experiment with a number of synthetic datasets that have been considered in previous studies. 
   The 2D ring dataset is a mixture of eight 2D Gaussians with mean $(\cos((2\pi/8)i), \sin((2\pi/8)i))$, and standard deviation of 0.01 in each dimension for $i \in \{1,\hdots,8\}$. We also experiment with a 2D grid dataset, which is a mixture of 25 and 36, 2D Gaussians with mean $(-1 + 0.4i, -1 + 0.4j)$ and standard deviation of 0.05 in each dimension for $i,j \in \{0,1,\hdots,n\}$ where $n$ is the root square of number of mixture modes. Examples of synthetic data are illustrated in Figure~\ref{fig:Synthetic_Data}.
\begin{figure}[!ht]
    \centering
    \includegraphics[width=1\textwidth]{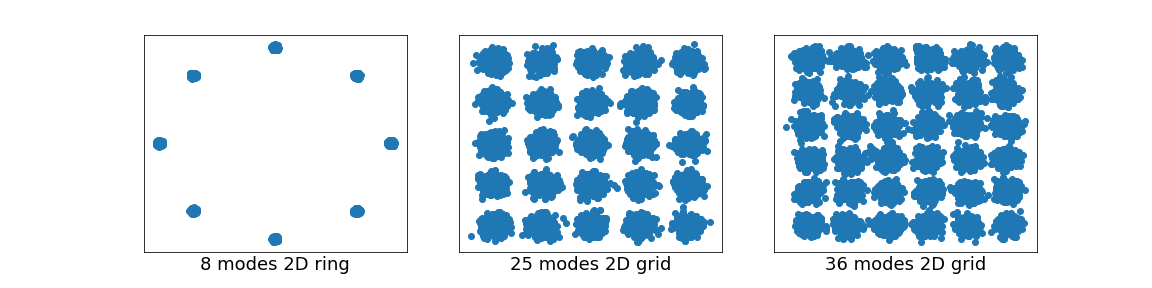}
    \caption{Synthetic data}
    \label{fig:Synthetic_Data}
\end{figure}

\item \emph{Stacked MNIST:}
Stacked MNIST data contains three channels, each presenting one digit from the MNIST dataset. Therefore, the stacked MNIST dataset covers 1000 different modes illustrated in Figure~\ref{fig:Stacked_MNIST_data}.
\begin{figure}[!ht]
    \centering
    \includegraphics[width=0.4\textwidth]{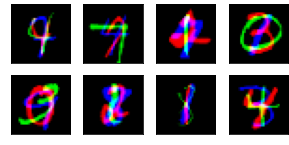}
\caption{Stacked MNIST data}
\label{fig:Stacked_MNIST_data}
\end{figure}

\item \emph{EES dataset:} 
Original EES dataset consists of full-wave electromagnetic simulations with the element shape represented by a binary image and its corresponding transmission response over frequency.
The EES element shapes are 2D binary images with 8-fold-symmetry where a single octant of the image contains sufficient information to fully describe the topology. 
The exhaustive search space includes 32,768 unique designs, which are all the possible combinations of the binary 15 bits, representing $\frac{1}{8}$ of the image. Similar dataset is generated for 19$\times$19 designs, though unlike the 9$\times$9 design dataset, it only contains a small subset of the corresponding search space.

A sample set of EES elements is shown in Figure~\ref{fig:EES_Dataset}. 
The high pass (HP) and low pass (LP) designs are the two main categories of EES designs. In the EES dataset, the high pass designs, unlike low pass designs, have edge to edge connection, which is illustrated in Figure~\ref{fig:HP}.
As mentioned above, the whole search space is available for 9$\times$9 designs; however, for 19$\times$19 designs, 100,000 samples form the respective dataset. The dataset used in this paper is synthetically generated based on the structural configurations of the original dataset considered in \citep{mohammadjafari2021designing}.

\begin{figure}[!ht]
\centering
\subfloat[High pass \label{fig:HP}]{\includegraphics[width=0.45\textwidth]{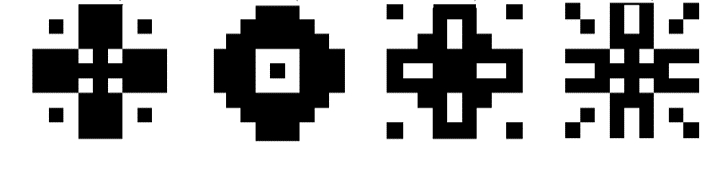}}
\hfill
\subfloat[Low pass \label{fig:LP}]{\includegraphics[width=0.45\textwidth]{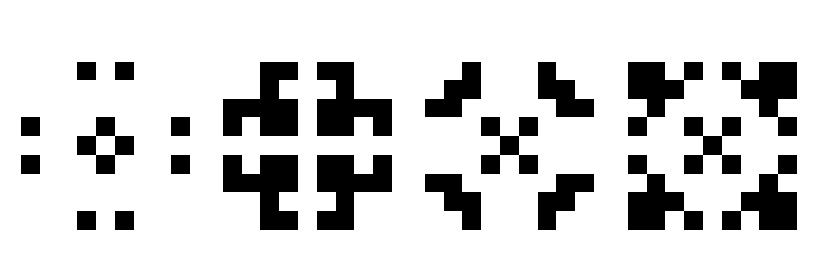}}
\caption{EES dataset}
\label{fig:EES_Dataset}
\end{figure}

\end{itemize}

\subsection{Experimental setup}
We evaluate our proposed VARGAN architecture on three different datasets and compare the performance with vanilla GAN and other popular GAN variants from the literature. Models include GANs with mini-batch discrimination (MB) \citep{salimans2016improved}, PacGAN \citep{lin2018pacgan} with packing factor of four, MADGAN \citep{ghosh2018multi} with four generators, GDPP \citep{elfeki2019gdpp} and PacVARGAN. 
PacVARGAN is a new model with a discriminator network incorporating the PacGAN structure with a packing factor of four and an additional VarNet network.
Although, GAN structures in our experiments are not conditional, in the second part of our experiments, we investigate the effect of conditioning on the mode collapse. 
We add the categories as labels to the best performing unconditioned model and compare the performance with baseline vanilla GAN models. 
Only EES dataset is used for conditional GAN experiments.

The feed-forward (FF) GAN architecture is similar for all the datasets, whereas the convolutional GAN structure is dataset dependent, and reported in Appendix~\ref{ap:model_arch} for stacked MNIST and EES datasets. 
For the synthetic 2D data, a feed-forward architecture is selected, since data does not benefit from a convolutional structure. 
On the other hand, for stacked MNIST and EES, both feed-forward and convolutional neural networks are employed. 

For synthetic and stacked MNIST data, each model is trained on 100,000 training samples and tested on 26,000 samples. For all the networks, Adam optimizer with a learning rate of $2e-4$ is used. For the EES data, each model is trained on 100,000 samples for $19\times19$ designs, and on all the designs in the search space for $9\times9$ designs. 
The models are evaluated on 20,000 samples per category. 
The model is trained for 50, 30 and 400 epochs for synthetic, stacked MNIST and EES data. 
In VARGAN, MCR parameter values, namely, $L$, $S_1$, $S_2$, and $C$ are selected as 1, 10, 5, and 1 across all datasets based on our hyperparameter tuning experiments (see Appendix~\ref{ap:Limited_data_formula}). 
In our numerical study, to minimize the random initialization effect, each experiment is repeated five times and average performance values are reported.
All the models were implemented using Python and PyTorch library, and experiments were performed on a NVIDIA GeForce RTX 2070 GPU with 8 GB of GPU RAM.

\section{Numerical results}\label{sec:results}
In this section, we first present performance metrics used to evaluate GAN architectures. 
Then, we report the performance of GAN architectures over a variety of datasets. 
Finally, we investigate the impact of conditioning on mode collapse in GANs.

\subsection{Performance metrics}
Evaluating mode collapse is a challenging task, except in cases such as synthetic data where the modes are explicitly defined. 
Therefore, in this work, each GAN architecture is evaluated based on a set of popular performance metrics relevant to the dataset specifications. 
Since the labels are available for each dataset, a pretrained classifier is used to determine the \textit{number of modes} generated by the models. 
The \textit{number of modes} and \textit{Kullback-Leibler (KL) divergence}, which measures the distance between the generated and real distributions are reported for each model over all the datasets. 
\textit{Percentage of high-quality samples} is another metric introduced in \citep{lin2018pacgan} for evaluating model performance over a synthetic dataset. 
It measures the proportion of generated samples that are closer than 3 standard deviations to the center of the mode. 
Finally, for the stacked MNIST data, we use the \textit{inception score} to evaluate the quality and diversity of the generated samples \citep{salimans2016improved}. 
It is important to note that the run time performance in each experiment is reported based on the model's training time in seconds, and does not reflect the theoretical complexity.

\subsection{Comparison of GAN architectures}
In this section, we report performance evaluation results for different GAN architectures over synthetic and real-world datasets. 
Along with our extensive comparative analysis, we also investigate the convergence behaviours of the models over training epochs in the training process. 

\subsubsection{Results with synthetic dataset}
We have reported the number of generated modes, KL divergence and percentage of high-quality samples for each synthetic dataset variants separately. 
As shown in Table~\ref{tbl:syntetic_results}, for 2D grids with 25 and 36 modes, VARGAN captures all the modes with low divergence metric while generating high quality samples, and requiring less training time.

\setlength{\tabcolsep}{6pt}
\renewcommand{\arraystretch}{1.3}
\begin{table}[!ht]
\caption{{\footnotesize Performance metrics for synthetic datasets averaged over 5 repeats {(all the GANs have FF structure, best performing model is bolded)}}}
\label{tbl:syntetic_results}
\begin{center}
\scalebox{0.81}{

\begin{tabular}{llllllll}
\toprule
\multicolumn{1}{c}{\multirow{3}{*}{\STAB{\rotatebox[origin=c]{90}{{Dataset}}}}} &
\multicolumn{1}{c}{\multirow{3}{*}{Model}} & &\multicolumn{3}{c}{Metrics} \\     
\cmidrule{4-6} 
\multicolumn{1}{c}{} & \multicolumn{1}{c}{} &        & \multirow{2}{*}{Modes} & \multirow{2}{*}{\begin{tabular}[c]{@{}l@{}}High quality  \\ samples\end{tabular}} & \multirow{2}{*}{\begin{tabular}[c]{@{}l@{}}KL \\ divergence\end{tabular}} & Time (sec)\\
\multicolumn{1}{c}{}   &   &         \\ 
\midrule
\multicolumn{1}{c}{\multirow{10}{*}{\STAB{\rotatebox[origin=c]{90}{{Ring 8 modes}}}}} & GAN
&  & 6.20 $\pm$ 1.72 & 0.87 $\pm$ 0.03 & 0.69 $\pm$ 0.24 & 2,731 $\pm$ 56\\

&GAN + MB 
&  & 4.00 $\pm$ 2.60 & 0.56 $\pm$ 0.30 & 0.80 $\pm$ 0.45 & 693 $\pm$ 26\\

&\B PacGAN & & \B 8.00 $\pm$ 0.00 & \B 0.71 $\pm$  0.09 & \B 0.03 $\pm$ 0.02 & \B 2,790 $\pm$ 6\\

&MAD-GAN 
& & 3.66 $\pm$ 2.42 & 0.25 $\pm$ 0.36 & 1.25 $\pm$ 0.57 & 1,393 $\pm$ 73\\

&GDPP 
& & 5.00 $\pm$  1.26 & 0.60 $\pm$ 0.20 & 1.18 $\pm$ 0.21 & 1,962 $\pm$ 182\\
\cmidrule{2-7}
&VARGAN && 4.80 $\pm$ 0.40 & 0.71 $\pm$ 0.23 & 1.05 $\pm$ 0.09 & 1,480 $\pm$ 31 \\
&PacVARGAN & & 8.00 $\pm$ 0.00 & 0.32 $\pm$ 0.03 & 0.17 $\pm$ 0.07 & 1,477 $\pm$ 16\\ 
\midrule
\multicolumn{1}{c}{\multirow{10}{*}{\STAB{\rotatebox[origin=c]{90}{{Grid 25 modes}}}}}
&GAN 
&& 25.00 $\pm$ 0.00 & 0.85 $\pm$ 0.04 & 0.45 $\pm$ 0.14 & 3,608 $\pm$ 124 \\
&GAN + MB 
&&  5.60 $\pm$ 9.77 & 0.76 $\pm$ 0.38 & 1.94 $\pm$ 1.55 & 962 $\pm$ 2\\
& PacGAN 
&&
 25.00 $\pm$ 0.00 &  0.66 $\pm$ 0.07 &  0.15 $\pm$ 0.04 &  3,642 $\pm$ 89\\

 &MAD-GAN 
 && 24.80 $\pm$ 0.40 & 0.63 $\pm$ 0.23 & 0.57 $\pm$ 0.89 & 1,679 $\pm$ 81 \\

 &GDPP 
 &&25.00 $\pm$ 0.00 & 0.67 $\pm$ 0.02 & 0.26 $\pm$ 0.04 &  2,463 $\pm$ 85\\
\cmidrule{2-7}
&\B VARGAN && \B 25.00 $\pm$ 0.00 & \B 0.80 $\pm$ 0.04 & \B 0.16 $\pm$ 0.06 & \B 814 $\pm$ 23\\
& PacVARGAN &&  25.00 $\pm$ 0.00 &  0.59 $\pm$ 0.02 &  0.15 $\pm$ 0.05 &  833 $\pm$ 3\\
\midrule  
\multicolumn{1}{c}{\multirow{10}{*}{\STAB{\rotatebox[origin=c]{90}{{Grid 36 modes}}}}}
&GAN 
&  &  36.00 $\pm$ 0.00 & 0.84 $\pm$ 0.06 & 0.18 $\pm$ 0.06 & 6,154 $\pm$ 201\\
&GAN + MB 
&&  21.80 $\pm$ 17.39 & 0.66 $\pm$ 0.34 & 0.75 $\pm$ 1.41 & 1,136 $\pm$ 4\\
& PacGAN 
& &  36.00 $\pm$ 0.00 & 0.72 $\pm$ 0.03 & 0.09 $\pm$ 0.03 & 6,183 $\pm$ 228\\

& MAD-GAN 
& & 32.00 $\pm$ 6.92 & 0.57 $\pm$ 0.28 & 0.39 $\pm$ 0.40 & 1,804 $\pm$ 3\\

& GDPP 
& & 35.80 $\pm$ 0.40 & 0.70 $\pm$ 0.02 & 0.13 $\pm$ 0.11 & 2,712 $\pm$ 306\\
\cmidrule{2-7}
& \B VARGAN &&  \B 36.00 $\pm$ 0.00 & \B 0.73 $\pm$ 0.02 & \B 0.05 $\pm$ 0.01 & \B 761 $\pm$ 6\\
& PacVARGAN & &  36.00 $\pm$ 0.00 & 0.66 $\pm$ 0.01 & 0.11 $\pm$ 0.04 &  774 $\pm$ 4\\
\bottomrule         
\end{tabular}
}
\end{center}
\end{table}

We note that VARGAN does not perform well for 2D ring with 8 modes, whereas PacGAN generates a high number of modes as well as high quality samples. 
Overall, it is observed that PacGAN consistently generates more unique samples with better quality for all three groups of synthetic data. 
However, the training times for PacGAN is much higher than other models, and is almost equal to the vanilla GAN model. 
Our results for PacGAN model on 2D ring and 2D grid with 25 modes are consistent with the findings in \citep{lin2018pacgan}.
By combining the PacGAN properties with VARGAN, PacVARGAN shows both high number of generated modes and low run times. 
We also find that mini-batch discrimination does not perform well for different designs. 
The reason can be the network's feed-forward architecture that cannot benefit from mini-batch discrimination approach in addressing the mode collapse. 
MADGAN outperforms mini-batch discrimination, but still does not perform as well as other models. 
Moreover, GDPP generates high number of unique designs that match the findings of \citet{elfeki2019gdpp}. 
We find that VARGAN outperforms GDPP in terms of percentage of high quality samples and KL divergence. 

\begin{figure}[!ht]
\centering
\subfloat[Number of modes \label{fig:36_modes}]{\includegraphics[width=0.85\textwidth]{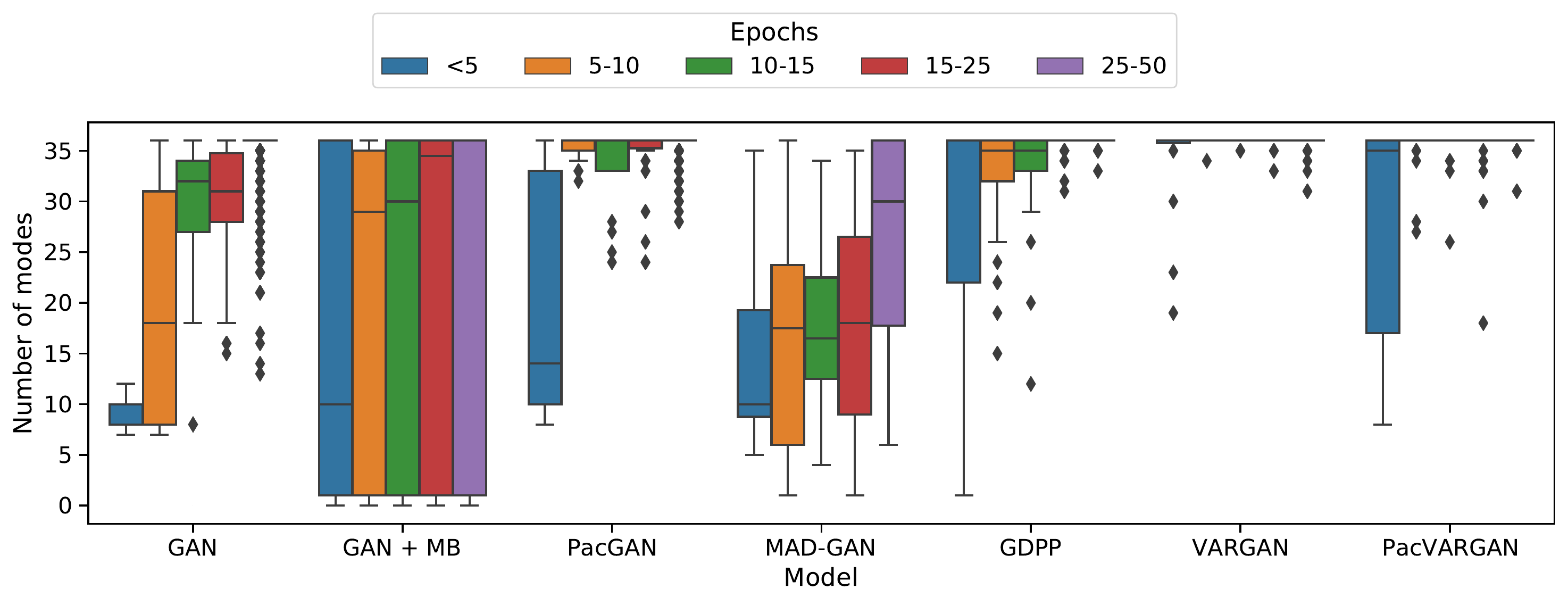}}
\vfill
\subfloat[KL divergence \label{fig:36_KL}]{\includegraphics[width=0.85\textwidth]{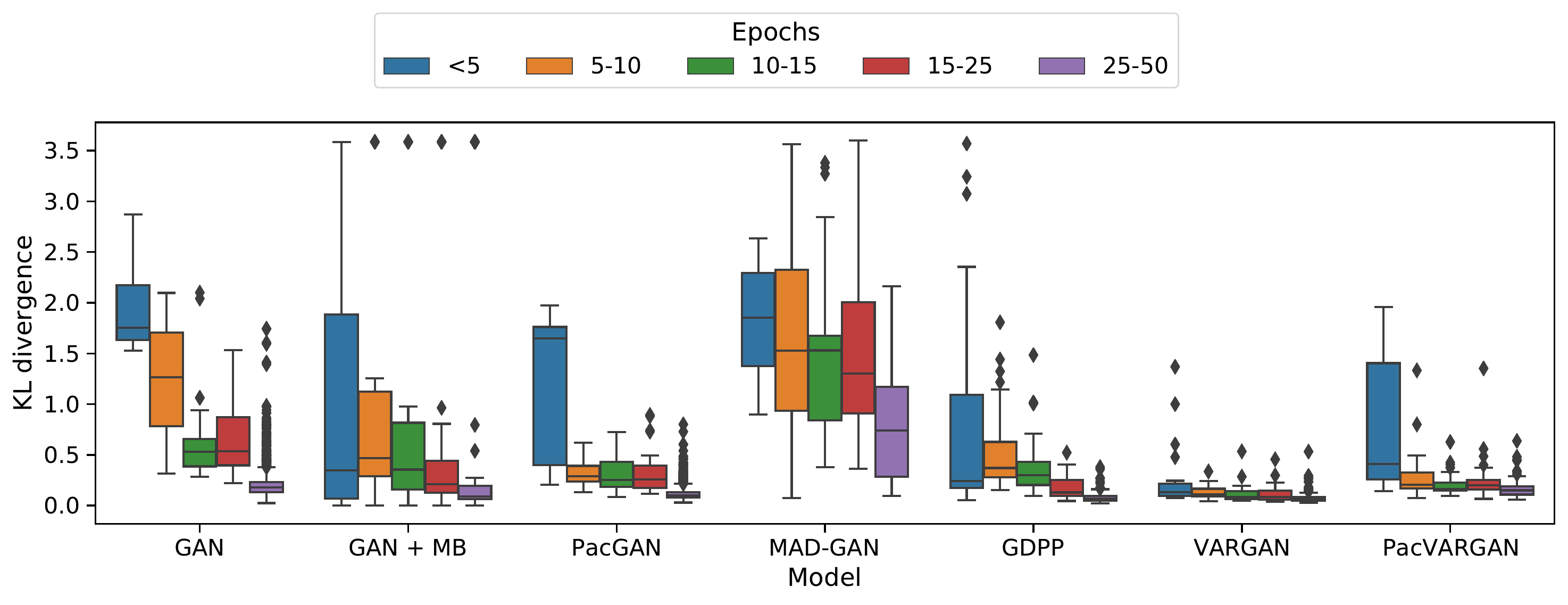}}
\vfill
\subfloat[Percentage of high quality samples \label{fig:36_highq}]{\includegraphics[width=0.85\textwidth]{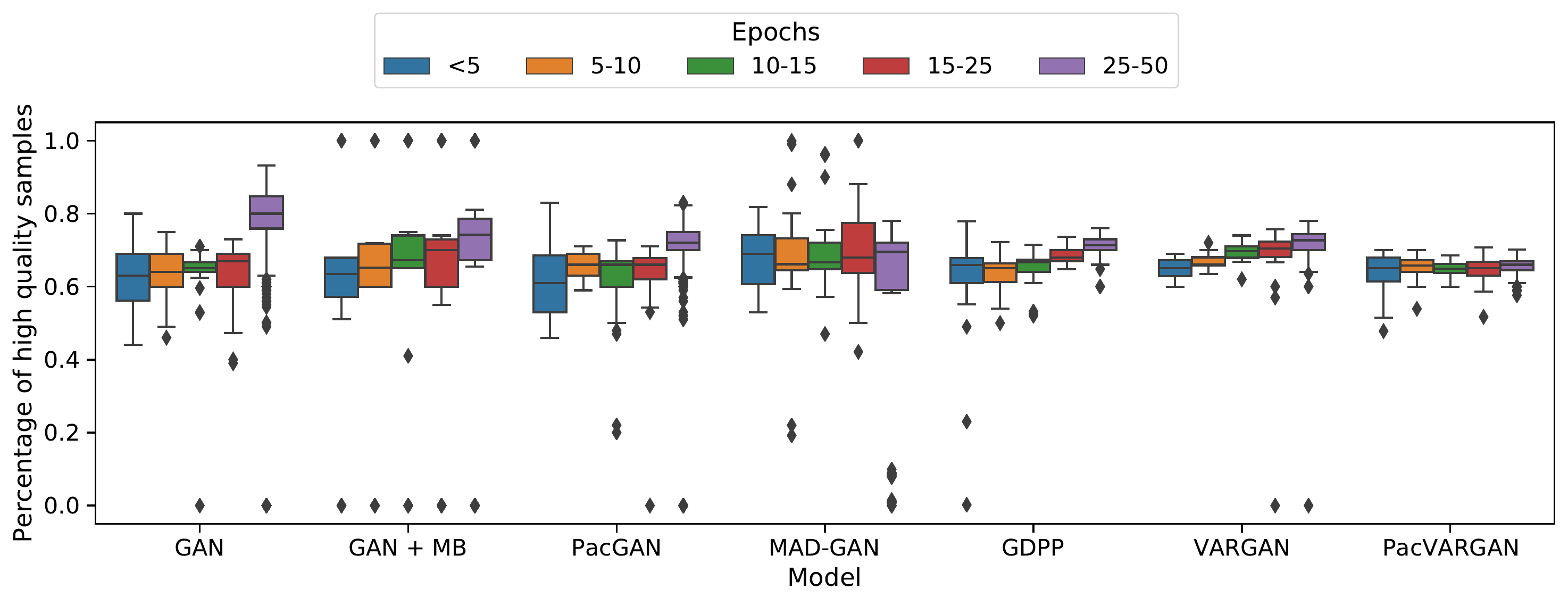}}
\caption{Comparison of different GAN models on synthetic 2D grid data with 36 modes (results are averaged over 5 repeats)}
\label{fig:Synthetic_result_36grid_boxplot}
\end{figure}

Further analysis over the 50 epoch training process for 2D ring data shows that VARGAN has a quicker convergence process, and leads to a better performance for 2D ring data at the initial epochs compared to the other models.
Result for 2D grid data with 36 modes is illustrated in Figure~\ref{fig:Synthetic_result_36grid_boxplot} and shows a great performance for the VARGAN in number of generated modes, as well as high quality samples. Although all the models except GAN+MB converge to target number of modes, the variance over the repeats points to a better performance for VARGAN. 
Note that we repeated this analysis with other two synthetic data variants, which lead to similar observations (see Appendix~\ref{ap:detailed_results} for more details).
Overall, these results largely confirm the fast convergence of VARGAN on earlier epochs of the training. 
Furthermore, they point to the methodological/architectural differences between different GAN variants.

\subsubsection{Results with stacked MNIST dataset}
Table~\ref{tbl:MNIST_result_last_epoch} summarizes the performance of convolutional (Conv) and feed-forward (FF) structures for different GAN models over stacked MNIST data. 
The results show that, overall, all the models perform well for both structures. 
FF (vanilla) GAN captures all the modes, and attains a high inception score and a low KL divergence. 
GDPP and VARGAN with FF structures also generate a high number of modes and a low KL divergence. 
Results with convolutional structures show VARGAN as the best performing model with high number of modes, low KL divergence and high inception score. 
Compared to its FF counterpart, VARGAN performance drops slightly with convolutional structure, however, run time improves significantly.
In contrast, PacVARGAN performance benefits significantly from the convolutional structure.
The results also confirm that the MB method fails to perform well with FF structure (similar to synthetic data), and its performance improve with convolutional structure. 
It is worth mentioning that FF PacGAN model performs slightly worse than its convolutional counterpart. 
One possible reason for this behaviour might be the PacGAN discriminator structure which combines $m$ samples, and changes the image data input in the FF model case. 
On other hand, in convolutional models, PACGAN channel factor gets multiplied by $m$ and image structure remains the same. 
\setlength{\tabcolsep}{6pt}
\renewcommand{\arraystretch}{1.3}
\begin{table}[!ht]
\caption{{\footnotesize Performance metrics for stacked MNIST dataset (with 1000 modes) averaged over 5 repeats {(best performing model is bolded)}}}
\label{tbl:MNIST_result_last_epoch}
\begin{center}
\scalebox{0.8}{
\begin{tabular}{lllllllll}
\toprule
\multicolumn{1}{c}{\multirow{3}{*}{\STAB{\rotatebox[origin=c]{90}{{Structure}}}}}& 
\multicolumn{1}{c}{\multirow{3}{*}{Models}} & &\multicolumn{3}{c}{Metrics} \\     
\cline{3-6} 
\multicolumn{1}{c}{} &&& \multirow{2}{*}{Modes} & \multirow{2}{*}{\begin{tabular}[c]{@{}l@{}}Inception \\ Score\end{tabular}} & \multirow{2}{*}{\begin{tabular}[c]{@{}l@{}}KL \\ divergence\end{tabular}} & Time (sec)\\
\multicolumn{1}{c}{}   &   &         \\ \midrule

\multicolumn{1}{c}{\multirow{7}{*}{\STAB{\rotatebox[origin=c]{90}{{FF}}}}}& \B GAN&& \B 1000 $\pm$ 0 & \B 1.74 $\pm$ 0.01 & \B 0.13 $\pm$ 0.00 & \B 981 $\pm$ 0\\

& GAN + MB&  &790 $\pm$	257 &	1.61 $\pm$	0.35 &	1.10 $\pm$	1.18 & 1,447 $\pm$ 2\\
&PacGAN && 920 $\pm$ 10 & 1.82 $\pm$ 0.05 & 0.79 $\pm$ 0.05 & 1,094 $\pm$ 9\\

&MAD-GAN && 994 $\pm$ 11 & 1.56 $\pm$ 0.02& 0.72 $\pm$ 0.12 & 1,589 $\pm$ 30\\

&GDPP && 995 $\pm$ 2 & 1.74 $\pm$ 0.04 & 0.40 $\pm$ 0.03 & 2,467 $\pm$ 14\\

\cmidrule{2-7}
&\B VARGAN && \B 999  $\pm$  0 & \B 1.73 $\pm$ 0.02 & \B 0.24 $\pm$ 0.00 & \B 4,686 $\pm$ 17
\\

&PacVARGAN && 794 $\pm$ 42 & 1.95 $\pm$ 0.10 & 1.19 $\pm$ 0.14 & 4,862 $\pm$ 1
\\
\midrule
\multicolumn{1}{c}{\multirow{7}{*}{\STAB{\rotatebox[origin=c]{90}{{Conv}}}}}
&GAN && 956 $\pm$ 23 & 1.88 $\pm$ 0.06 & 0.39 $\pm$ 0.11 & 1,203 $\pm$ 2
\\
&  GAN + MB  && 916 $\pm$ 54 & 2.09 $\pm$ 0.09 & 0.76 $\pm$ 0.19 & 1,978 $\pm$ 0 
\\

&  PacGAN  && 935 $\pm$ 20 & 2.09 $\pm$ 0.11 & 0.84 $\pm$ 0.13 & 1,981 $\pm$ 1
\\
&MAD-GAN && 959 $\pm$ 11 & 1.99 $\pm$ 0.05 & 0.87 $\pm$ 0.11 & 5,532 $\pm$ 7\\

&GDPP && 943 $\pm$ 52 & 2.18 $\pm$ 0.07 & 0.69 $\pm$ 0.23 & 3,334 $\pm$ 20\\

\cmidrule{2-7} 
&\B VARGAN &&  \B 997 $\pm$ 1 & \B 2.16 $\pm$ 0.02 & \B 0.34 $\pm$ 0.06 & \B 2,336 $\pm$ 11
\\
& \B PacVARGAN 
&&  \B 996 $\pm$ 2 & \B 2.14 $\pm$ 0.04 & \B 0.30$\pm$ 0.03 & \B 2,243 $\pm$ 3\\
\bottomrule
\end{tabular}

}
\end{center}
\end{table}

We note that the number of uniquely generated modes for convolutional PacGAN, and GDPP models matches the results in \citep{lin2018pacgan,elfeki2019gdpp} studies. 
We also examine the models' performance over the training epochs using GANs with convolutional structures (see Appendix~\ref{ap:detailed_results}). 
We find that VARGAN and PacVARGAN show fast early convergence in terms of all three metrics, which makes them great candidates for artificial data generation. 
In addition, earlier convergence can be beneficial in terms of speed of GAN training.

\subsubsection{Results with EES dataset}\label{sec:EES_results}
Table~\ref{tbl:EES_result_last_epoch_9} shows the model performance for feed-forward and convolutional structures over 9$\times$9 and 19$\times$19  EES designs.
Overall, results indicate that convolutional models generally have a better performance compared to feed-forward models.
For $9\times 9 $ designs, for the FF structures, VARGAN, GDPP and GAN create high number of unique modes as well as achieving reasonable accuracy with GDPP achieving the best KL divergence.
On the other hand, training time is significantly higher for GDPP compared to GAN and VARGAN.
PacGAN performs poorly for FF structures for both 9$\times$9 and 19$\times$19 designs, similar to stacked MNIST.
MADGAN with five generators does not perform well for FF structure, and completely collapse and generate one high quality sample with this structure.
\setlength{\tabcolsep}{2.7pt}
\renewcommand{\arraystretch}{1.3}
\begin{table}[!ht]
\caption{{\footnotesize Performance metrics for feed-forward and convolutional structures on 9$\times$9 and 19$\times$19 EES designs with 2 categories averaged over 5 repeats {\footnotesize {(Best performing model is bolded)}}}}
\label{tbl:EES_result_last_epoch_9}
\begin{center}
\scalebox{0.69}{

\begin{tabular}{lllllcccll}
\toprule
\multicolumn{1}{c}{\multirow{3}{*}{\STAB{\rotatebox[origin=c]{90}{{Design}}}}} & 
\multicolumn{1}{c}{\multirow{3}{*}{\STAB{\rotatebox[origin=c]{90}{{Structure}}}}} &
\multicolumn{1}{c}{\multirow{3}{*}{Models}} & \multicolumn{2}{c}{Modes} && \multicolumn{2}{c}{Accuracy} & & \\     
\cmidrule{4-5} \cmidrule{7-8} 

\multicolumn{1}{c}{} &  & &  \multirow{2}{*}{HP} & \multirow{2}{*}{LP} && \multirow{2}{*}{HP} & \multirow{2}{*}{LP} & \multirow{2}{*}{\begin{tabular}[c]{@{}l@{}}KL \\ divergence\end{tabular}} & Time (sec)\\
\multicolumn{1}{c}{}   &   &         \\ \midrule
\multicolumn{1}{c}{\multirow{14}{*}{\STAB{\rotatebox[origin=c]{90}{{$9\times9$}}}}}&
\multicolumn{1}{c}{\multirow{7}{*}{\STAB{\rotatebox[origin=c]{90}{FF}}}} & GAN &  14,572 $\pm$ 94 & 6,653 $\pm$ 112 &&  0.3121 $\pm$ 0.0150 & 0.6774 $\pm$ 0.0164 & 0.00047 $\pm$ 0.00067 & 677 $\pm$ 9\\
& &GAN + MB &    11,654 $\pm$ 5,828  & 5,071 $\pm$  2,534 && 0.4284 $\pm$  0.2480 & 0.5707 $\pm$ 0.2505 & 0.17226 $\pm$ 0.34417 & 672 $\pm$ 12\\	

&& PacGAN & 2,603 $\pm$  757 & 1,148 $\pm$  303    &&  0.2970 $\pm$  0.0181 & 0.7078 $\pm$  0.0180 & 0.00066 $\pm$   0.00026 & 663 $\pm$ 1\\

&&MAD-GAN  & - & - && - & - & -\\

&&\B GDPP &  \B 14,580 $\pm$ 175 & \B 6,577 $\pm$ 191  && \B 0.3118 $\pm$ 0.0131 & \B 0.6862 $\pm$ 0.0132 & \B 0.00026 $\pm$ 0.00042 & \B 2,969 $\pm$ 19 \\
\cmidrule{3-10}

&&\B VARGAN & \B 14,793 $\pm$ 137 &  \B 6,388 $\pm$ 259  &&  \B 0.2945 $\pm$ 0.0128 & \B 0.7015 $\pm$ 0.0166 & \B 0.00060 $\pm$ 0.00054 & \B 996 $\pm$ 13\\
&&PacVARGAN &  1,939 $\pm$ 1,070 & 826 $\pm$  431   && 	0.2716 $\pm$ 0.0615 & 0.7264 $\pm$ 0.0497 & 0.01053 $\pm$ 0.01516 & 929 $\pm$ 4\\	
\cmidrule{2-10}

&\multicolumn{1}{c}{\multirow{7}{*}{\STAB{\rotatebox[origin=c]{90}{{Conv}}}}} &
GAN &  14,928 $\pm$ 	107 &6,592 $\pm$ 159	&& 	0.1822 $\pm$ 	0.0150	&  0.4035 $\pm$ 	0.0177 & 	0.00040 $\pm$ 	0.00020	& 3,704 $\pm$ 1,083\\
&&\B GAN + MB  & \B 15,009 $\pm$ 118 & \B 6,716 $\pm$ 95 	&& \B 0.1727 $\pm$ 	0.0011 	& \B 0.3739	$\pm$ 0.0148 & \B	0.00010 $\pm$ 0.00008 & \B 18,133 $\pm$ 1,805\\	
&&  PacGAN  &	 14,881 $\pm$ 	146	&  6,714	 $\pm$ 210 && 	 0.2315 $\pm$ 	0.0168	& 	 0.4976 $\pm$ 	0.0264	& 	0.00062 $\pm$ 	0.00029 &  1,384 $\pm$ 32	\\

&&MAD-GAN  & 359 $\pm$  469 & 144 $\pm$  135  && 0.0431 $\pm$  0.0773 & 0.0301 $\pm$  0.0496 & 0.43074 $\pm$  0.34093 & 2,414 $\pm$ 20\\

&&GDPP & 11,899 $\pm$ 5,892 & 5,303 $\pm$ 2,602  && 0.1487 $\pm$ 0.0267 & 0.2916  $\pm$ 0.1264 & 0.00292 $\pm$ 0.00579 & 3,036 $\pm$ 66\\
\cmidrule{3-10}
&& \B VARGAN & 	\B 14,962 $\pm$ 164 & \B 6,659 $\pm$ 123	&& \B 0.1608 $\pm$ 	0.0218 & \B	0.3537 $\pm$ 0.0448	& \B 0.00014 $\pm$ 0.00015 & \B 1,841 $\pm$ 487\\
&&PacVARGAN &		14,829 $\pm$ 	224 & 6,729 $\pm$ 	179		&&  0.2131 $\pm$ 0.0065	& 	0.4567 $\pm$ 	0.0207	& 	0.00054 $\pm$ 	0.00059 & 3,072 $\pm$ 1,130\\	
\midrule
\midrule
\multicolumn{1}{c}{\multirow{14}{*}{ \STAB{\rotatebox[origin=c]{90}{{$19\times19$}}}}}&\multicolumn{1}{c}{\multirow{7}{*}{\STAB{\rotatebox[origin=c]{90}{{FF}}}}} &
\B GAN & \B 13,148 $\pm$ 7,349 & \B 6,351 $\pm$ 3,135  && \B 0.3374 $\pm$ 0.0911 & \B 0.6576 $\pm$ 0.0942 & \B 0.07343 $\pm$ 0.05317 & \B 439 $\pm$ 12\\
&&GAN + MB & 9,698 $\pm$  5,824 &  4,443 $\pm$ 2,139 &&  0.3621 $\pm$ 0.1208 & 0.6260 $\pm$ 0.1191 & 0.06655 $\pm$ 0.03573 & 448 $\pm$ 29\\	

&&PacGAN  & 2,232 $\pm$ 1,271 & 1,078 $\pm$ 741 && 0.4046 $\pm$ 0.3167 & 0.5928 $\pm$ 0.3152 & 0.26998 $\pm$  0.25194 & 369 $\pm$ 30\\

&&MAD-GAN  &- & - && - & - & -\\
&&GDPP & 5,232 $\pm$ 5,206 & 2,625 $\pm$ 1,196 && 0.4412 $\pm$ 0.2078 & 0.5669 $\pm$ 0.2069 & 0.09441 $\pm$ 0.07688 & 1,241 $\pm$ 38\\
\cmidrule{3-10}

&&\B VARGAN & \B 13,534 $\pm$ 6,718 & \B 5,204 $\pm$ 2,954 && \B 0.2784 $\pm$ 0.0917 & \B 0.7155 $\pm$ 0.0895 & \B 0.11176 $\pm$ 0.06120 & \B 673 $\pm$ 52\\
&&PacVARGAN  &  11,266 $\pm$ 3,077 & 4,476 $\pm$ 1,696 && 0.2810 $\pm$ 0.0426 &  0.7153 $\pm$ 0.0443 &  0.10882 $\pm$ 0.03999 & 637 $\pm$ 21	\\	
\cmidrule{2-10}
&\multicolumn{1}{c}{\multirow{7}{*}{\STAB{\rotatebox[origin=c]{90}{{Conv}}}}} &
GAN &  237 $\pm$ 229 & 258 $\pm$ 346 && 0.6799 $\pm$ 0.2050 & 0.3216 $\pm$ 0.2050 & 0.17950 $\pm$ 0.22395 & 975 $\pm$ 60	\\

&&GAN + MB & 521 $\pm$ 611 & 550 $\pm$ 670 && 0.6138 $\pm$ 0.2615 & 0.3839 $\pm$ 0.2600  &0.20468 $\pm$ 0.25432 & 3,766 $\pm$ 60\\

&& \B PacGAN  & \B 15,343  $\pm$  1,180  & \B 7,260  $\pm$  1,714&& \B 0.6805  $\pm$ 0.0442 & \B 0.3193  $\pm$ 0.0471 & \B 0.07148 $\pm$  0.03324 & \B 908 $\pm$ 11\\
&&MAD-GAN  & 3,166 $\pm$  2,037 &  689 $\pm$  860  && 0.0275 $\pm$  0.0477 & 0.1431 $\pm$  0.1108 & 0.26012 $\pm$  0.08118 & 2,256 $\pm$ 39\\
&&GDPP  &  501 $\pm$ 451 & 271 $\pm$ 329 && 0.7499 $\pm$ 0.3189 & 0.2509 $\pm$ 0.3192 & 0.42096 $\pm$ 0.24309 & 1,562 $\pm$ 73 \\
\cmidrule{3-10}
 &&VARGAN & 1,848 $\pm$ 1,316  & 896 $\pm$ 415 && 0.6114  $\pm$  0.2360 & 0.3878 $\pm$ 0.2379 & 0.15896  $\pm$ 0.18456 & 578 $\pm$  50\\

 &&\B PacVARGAN & \B 13,965 $\pm$  1,039 & \B 5,318  $\pm$  1,606 && \B 0.7399  $\pm$ 0.0672 &
\B 0.2574  $\pm$  0.0696 & \B 0.13430 $\pm$  0.08211 &  \B 467 $\pm$  50\\

\bottomrule
\end{tabular}

}
\end{center}
\end{table}

In convolutional models, GAN, GAN+MB, VARGAN and PacVARGAN generate the highest number of high pass designs. 
Lowest KL divergence values are also achieved by GAN+MB.
However, the training time for GAN+MB is excessively high, and approximately 10 times higher than VARGAN model for the $9\times9$ designs. 


Intuitively, we expect the performance to deteriorate for the convolutional models in $19\times19$ designs compared to $9\times 9$ designs. 
The input in this case is a two dimensional matrix with two symmetrical parts, and the model should figure out its symmetry to generate acceptable designs. 
This creates a more difficult task compared to the feed-forward models and $9\times9$ design generation. 
It is important to note that the entire design search space is not available in $19\times19$ case. 
We observe that all the models outperform the vanilla GAN with high number of uniquely generated modes. 
Moreover, PacGAN, VARGAN and PacVARGAN report the lowest KL divergence compared to the rest of the models.
Overall, for $19\times19$ designs, best performance is achieved by PACGAN with convolutional structures, followed up by PacVARGAN. 
In general, we observe more robustness for the convolutional structures as evidenced by lower standard deviation over a variety of performance metrics.
We also find that the convergence of models over training epochs does not show any significant superiority for VARGAN compared to other GAN models for the EES dataset. 

\subsection{Impact of conditioning on mode collapse}
\citet{mao2019mode} argue that conditioning distracts the focus of generator from input noise, which is responsible to create variant designs. 
As a result, the generator is encouraged to create deterministic outputs based on the conditioned vector, causing mode collapse. 
To investigate the effect of conditioning on mode collapse, we have selected the best performing models from Section~\ref{sec:EES_results} to experiment with their conditional version. 
Results for this experiment are reported in Table~\ref{tbl:EES_conditioned}.

\setlength{\tabcolsep}{2.5pt}
\renewcommand{\arraystretch}{1.3}
\begin{table}[!ht]
\caption{{\footnotesize Comparison of conditioned and unconditioned models on EES designs {\footnotesize {(bolded models present the best performing model among the conditioned ones)}}}}
\label{tbl:EES_conditioned}
\begin{center}
\scalebox{0.73}{

\begin{tabular}{lccccccccccc}
\toprule
\multicolumn{1}{c}{\multirow{3}{*}{\STAB{\rotatebox[origin=c]{90}{{Design}}}}}&
\multicolumn{1}{c}{\multirow{3}{*}{\STAB{\rotatebox[origin=c]{90}{{Structure}}}}}&
\multicolumn{1}{c}{\multirow{3}{*}{\STAB{\rotatebox[origin=c]{90}{{Models}}}}} &
\multicolumn{1}{c}{\multirow{3}{*}{\STAB{\rotatebox[origin=c]{90}{{Condition}}}}}&
\multicolumn{2}{c}{Modes} && \multicolumn{2}{c}{Accuracy} &  \\     
\cmidrule{5-6} \cmidrule{8-9} 

\multicolumn{1}{c}{} &&&& \multirow{2}{*}{HP} & \multirow{2}{*}{LP} && \multirow{2}{*}{HP} & \multirow{2}{*}{LP} & \multirow{2}{*}{\begin{tabular}[c]{@{}l@{}}KL \\ divergence\end{tabular}} \\
\multicolumn{1}{c}{}   &   &         \\ \midrule
\multicolumn{1}{c}{\multirow{10}{*}{\STAB{\rotatebox[origin=c]{90}{{9$\times$9}}}}}& \multicolumn{1}{c}{\multirow{5}{*}{\STAB{\rotatebox[origin=c]{90}{{FF}}}}}  
 & GDPP &  No &   14,580 $\pm$ 175 & 6,577 $\pm$ 191  && 0.3118 $\pm$ 0.0131 & 0.6862 $\pm$ 0.0132 & 0.00026 $\pm$ 0.00042 \\
&  & GDPP &  Yes & 11,441 $\pm$  198 & 6,146 $\pm$ 122  && 0.9366 $\pm$  0.0072 & 0.9778 $\pm$  0.0073 & 0.06611 $\pm$  0.00292 \\

&  & \B MSGAN  & \B Yes &   \B 11,513 $\pm$  252 & \B 6,701 $\pm$  200  && \B 0.9556 $\pm$  0.0091 & \B 0.9820 $\pm$  0.0053 & \B 0.07120 $\pm$  0.00339 \\

&  & VARGAN  & No & 14,793 $\pm$ 137  &  6,388 $\pm$ 259 &&  0.2945 $\pm$ 0.0128 & 0.7015 $\pm$ 0.0166 & 0.00060 $\pm$ 0.00054\\
&  & VARGAN  & Yes & 11,338 $\pm$  364 & 6,083 $\pm$  321  && 0.9576 $\pm$  0.0062 & 0.9789 $\pm$  0.0030 & 0.07266 $\pm$  0.00307 \\

\cmidrule{2-10} &
\multicolumn{1}{c}{\multirow{7}{*}{ \STAB{\rotatebox[origin=c]{90}{{Conv}}}}} & 
PacGAN & No & 	 14,881 $\pm$ 	146 & 6,714	 $\pm$ 210	&& 	0.2315 $\pm$ 	0.0168	& 	0.4976 $\pm$ 	0.0264	& 	0.00062 $\pm$ 	0.00029	\\
&  & PacGAN & Yes & 11,572 $\pm$  3,856  & 5,585 $\pm$  1,917 && 0.5229 $\pm$  0.0361 & 0.7368 $\pm$  0.0893 & 0.01101 $\pm$ 0.00637 \\
&  & \B MSGAN & \B Yes &  \B 14,962 $\pm$  135 & \B 7,280 $\pm$  72   && \B 0.5672 $\pm$  0.0185 & \B 0.8410 $\pm$  0.0168 & \B 0.00660 $\pm$  0.00225 \\
&  & VARGAN  & No &  	14,962	 $\pm$ 164	& 6,659 $\pm$ 	123	&& 	0.1608 $\pm$ 	0.0218 	& 	0.3537 $\pm$ 	0.0448	& 	0.00014 $\pm$ 	0.00015\\
&  & VARGAN  & Yes &  4,927 $\pm$  6,264 &  2,217 $\pm$  2,851  && 0.4067 $\pm$  0.3674 & 0.4236 $\pm$  0.3543 & 0.10876 $\pm$  0.13333 \\
&  & PacVARGAN  & No & 14,829 $\pm$ 	224 & 6,729 $\pm$ 	179		&&  0.2131 $\pm$ 0.0065	& 	0.4567 $\pm$ 	0.0207	& 	0.00054 $\pm$ 	0.00059 \\
&  & PacVARGAN  & Yes & 9,207 $\pm$ 5,348 & 4,191 $\pm$ 2,581 && 0.4111 $\pm$ 0.1828 & 0.7138 $\pm$ 0.1223 & 0.00792 $\pm$ 0.00647 \\
\midrule 
\midrule
\multicolumn{1}{c}{\multirow{10}{*}{ \STAB{\rotatebox[origin=c]{90}{{19$\times$19}}}}} & \multicolumn{1}{c}{\multirow{5}{*}{\STAB{\rotatebox[origin=c]{90}{{FF}}}}} &   GAN &  No &  13,148 $\pm$ 7,349 & 6,351 $\pm$ 3,135  &&  0.3374 $\pm$ 0.0911 & 0.6576 $\pm$ 0.0942 & 0.07343 $\pm$ 0.05317\\

&  & GAN &  Yes  & 4,712 $\pm$  6,440 & 1,356 $\pm$  1,472 && 0.1264 $\pm$  0.0649 & 0.8762 $\pm$  0.0641 & 0.34576 $\pm$ 0.17531\\

&  & \B MSGAN   & \B Yes & \B 6,634 $\pm$  8,488 & \B 3,639 $\pm$  4,469 && \B 0.2830 $\pm$  0.1837 & \B 0.7371 $\pm$  0.1644 & \B 0.20766 $\pm$  0.26400\\
 &  &VARGAN  & No &   13,534 $\pm$ 6,718 & 5,204 $\pm$ 2,954 && 0.2784 $\pm$ 0.0917 & 0.7155 $\pm$ 0.0895 & 0.11176 $\pm$ 0.06120\\
&  &VARGAN  & Yes & 7,628 $\pm$  9,566 & 2,896 $\pm$  3,541 && 0.1296 $\pm$  0.1506 & 0.6807 $\pm$  0.3634 & 0.43210 $\pm$  0.28316 \\

\cmidrule{2-10} & \multicolumn{1}{c}{\multirow{7}{*}{ \STAB{\rotatebox[origin=c]{90}{{Conv}}}}} & 
PacGAN & No &   15,343  $\pm$  1,180  & 7,260  $\pm$  1,714&& 0.6805  $\pm$ 0.0442 & 0.3193  $\pm$ 0.0471 &  0.07148 $\pm$  0.03324\\
&  &\B PacGAN & \B Yes &  \B 13,393 $\pm$  8,004  & \B 4,361 $\pm$ 2,720 && \B 0.6765 $\pm$  0.1081 &\B  0.1165 $\pm$  0.0771 & \B 0.19064 $\pm$  0.12417   \\

&  &MSGAN   & Yes &  2,450 $\pm$  2,747 & 1,353 $\pm$  1,462 && 0.7131 $\pm$  0.2232 & 0.2592 $\pm$  0.2164 & 0.24830 $\pm$  0.25369\\
&  &VARGAN  & No &  1,848 $\pm$ 1,316 & 896 $\pm$ 415  && 0.6114  $\pm$  0.2360 & 0.3878 $\pm$ 0.2379 & 0.15896  $\pm$ 0.18456 \\
 &  &VARGAN  & Yes & 1,775 $\pm$  637 & 751  $\pm$ 407  && 0.6184 $\pm$  0.1575 &
 0.2080 $\pm$ 0.1580 & 0.12574 $\pm$  0.11315 \\
&  &PacVARGAN  & No & 13,965 $\pm$  1,039 & 5,318  $\pm$  1,606 && 0.7399  $\pm$ 0.0672 &
0.2574  $\pm$  0.0696 & 0.13430  $\pm$  0.08211\\

&  &PacVARGAN  & Yes& 1,126 $\pm$  776 &
675  $\pm$  561 && 0.7705 $\pm$  0.2262 & 0.1176  $\pm$  0.1161 &  0.32746  $\pm$ 0.24215\\
\bottomrule
\end{tabular}

}
\end{center}
\end{table}

We observe that, for all the models, conditioning reduces the number of uniquely generated modes over both designs. 
\citet{mao2019mode}'s MSGAN for conditional GANs outperforms all the conditioned versions of the models, except the PacGAN convolutional version for $19\times19$ designs.
We note that the conditional version of VARGAN for the feed-forward models outperforms the MSGAN in both uniquely generated modes and KL divergence. 
Moreover, conditional versions of VARGAN and PacVARGAN consistently lead to worse performance compared to their unconditional versions, indicating that there is no need to use a conditioned model to create reasonable number of samples for each category/mode for VARGAN variants.

\paragraph{\textbf{EES data with 8 categories}}
EES designs of size $9\times9$ can be further categorized to have eight classes \citep{mohammadjafari2021designing}. 
We next compare the effect of increasing the number of conditioned categories on diversity of generated samples. Specifically,
we use best performing unconditioned models and condition them on two and eight categories. Table~\ref{tbl:EES_8_Cat} presents the results for convolutional GAN, PacGAN, MSGAN and VARGAN variants for eight categories.

\setlength{\tabcolsep}{2.8pt}
\renewcommand{\arraystretch}{1.3}
\begin{table}[!ht]
\caption{{\footnotesize Performance metrics reported for $9\times9$ EES designs using convolutional models for 8 categories averaged over 5 repeats (HP 1: Band Pass, HP 2: Band Pass Stop, HP 3: High Pass, HP 4: True High Pass, HP 5: Very High Pass, LP 1: Band Stop, LP 2: Low Pass, LP 3: Stop Pass)}}
\label{tbl:EES_8_Cat}
\begin{center}
\scalebox{0.73}{

\begin{tabular}{cccccccc}
\toprule
\multicolumn{2}{c}{\STAB{\rotatebox[origin=c]{90}{{Category}}}} & GAN & MSGAN  & PacGAN & VARGAN & PacVARGAN & \multicolumn{1}{c}{\STAB{\rotatebox[origin=c]{90}{{Total samples}}}} \\ \midrule
\multicolumn{1}{c}{\multirow{6}{*}{\STAB{\rotatebox[origin=c]{90}{{HP}}}}} & 1   &  2,521 $\pm$ 238 & 2,664 $\pm$ 35 & 2,740 $\pm$ 15 & 1,543 $\pm$ 1,170 & 2,605 $\pm$ 279 & 2,804\\
\multicolumn{1}{c}{} & 2& 2,410 $\pm$ 116 & 2,531 $\pm$ 20 & 2,566 $\pm$ 12 & 1,452 $\pm$ 1,116 & 2,479 $\pm$ 166 &  2,614  \\
\multicolumn{1}{c}{}& 3& 2,945 $\pm$ 169 & 3,035 $\pm$ 45 &  3,130 $\pm$ 14 & 1,730 $\pm$ 1,323 & 2,972 $\pm$ 317 & 3,206\\
\multicolumn{1}{c}{} & 4& 7,819 $\pm$ 446 & 8,322 $\pm$ 27 & 8,359 $\pm$ 25 & 4,702 $\pm$ 3,665 & 8,011 $\pm$ 613 & 8,556\\
\multicolumn{1}{c}{}& 5& 5,093 $\pm$ 311 & 5,344 $\pm$ 28 & 5,399 $\pm$ 23 & 2,933 $\pm$ 2,350 & 5,101 $\pm$ 622 & 5,537\\ 
\cmidrule{2-8} 
\multicolumn{1}{c}{} & SUM &  20,788 & 21,896 & 22,194 & 12,364 & 21,168 & 22,717  \\ 
\midrule
\midrule
\multicolumn{1}{c}{\multirow{4}{*}{\STAB{\rotatebox[origin=c]{90}{{LP}}}}} & 1 & 3,788 $\pm$ 231 & 3,994 $\pm$ 45 & 4,026 $\pm$ 32 & 2,216 $\pm$ 1,758 & 3,816 $\pm$ 373 & 4,193  \\
\multicolumn{1}{c}{} & 2 & 4,083 $\pm$ 266 & 4,353 $\pm$ 17 & 4,370 $\pm$ 24 & 2,465 $\pm$ 1,959 & 4,090 $\pm$ 537 & 4,467\\
\multicolumn{1}{c}{} & 3   & 1,278 $\pm$ 73 & 1,342 $\pm$ 5 & 1,366 $\pm$ 6 & 761 $\pm$ 581 & 1,265 $\pm$ 182 & 1,391\\ 
\cmidrule{2-8}
\multicolumn{1}{c}{} & SUM & 9,149 & 9,689 & 9,762 & 5,442 & 9,171 & 10,051 \\
\midrule                  
& KL & 0.00407 $\pm$ 0.00283 & 0.00337 $\pm$ 0.00183 & 0.00156 $\pm$ 0.00100 & 0.08914 $\pm$ 0.10644 & 0.00394 $\pm$  0.00431\\   
\bottomrule
\end{tabular}
}
\end{center}
\end{table}

\begin{figure}[!ht]
\centering
\subfloat[{ High pass designs unique modes (right: 8 category data, left: 2 category data)} \label{fig:}]{\includegraphics[width=0.9\textwidth]{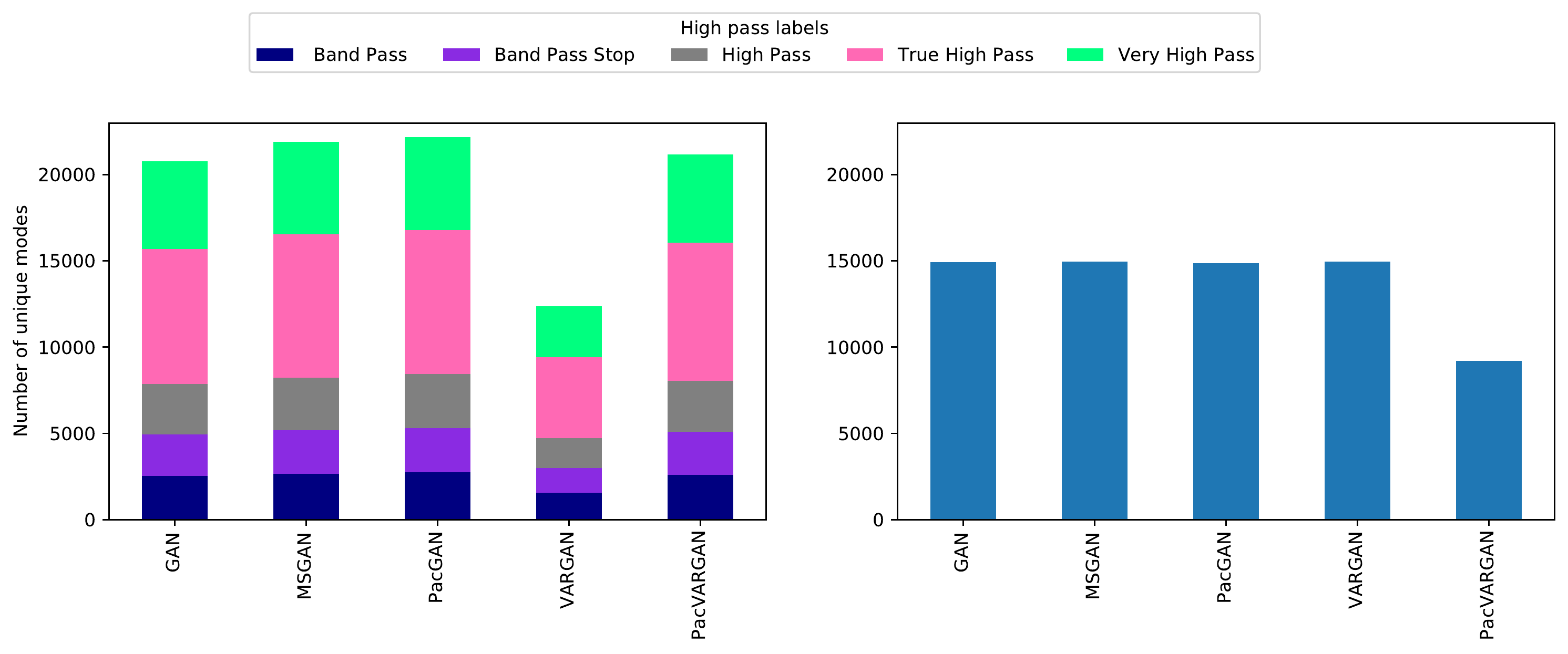}}
\vfill
\subfloat[{ Low pass designs unique modes (right: 8 category data, left: 2 category data)} \label{fig:}]{\includegraphics[width=0.9\textwidth]{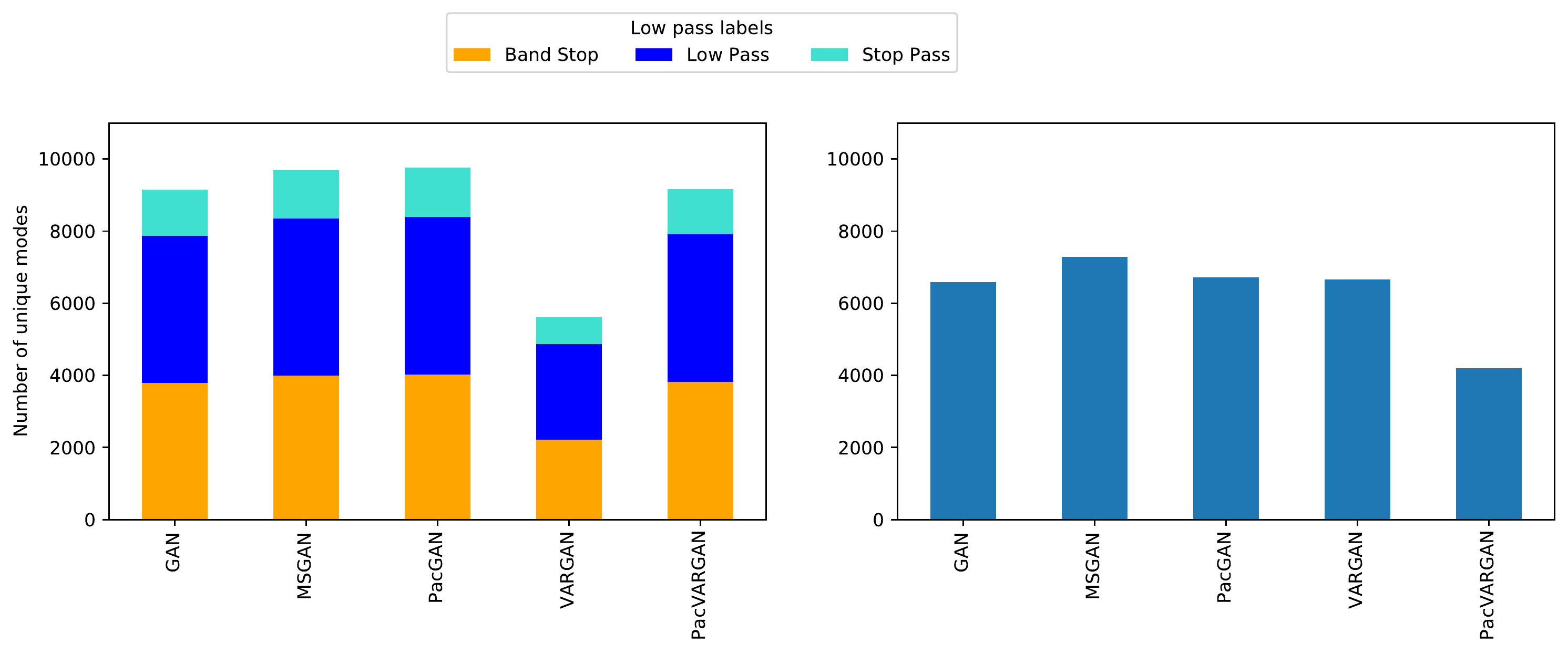}}
\caption{Comparison of total generated unique modes for high pass and low pass designs for 2 and 8 categories over $9\times9$ designs}
\label{fig:8_vs_2}
\end{figure}

As expected, PacGAN, MSGAN and PacVARGAN lead to a high performance in this case. 
Figure~\ref{fig:8_vs_2} provides a comparison of number of unique designs generated for 2 category and 8 category EES datasets.
The results show an increase in number of covered modes as the number of conditioned categories increases. 
The reason behind this result might be the one hot encoding structure of the labels. 
The large length of labels increases the randomness and diversity of the samples and encourages the generator to avoid creating deterministic results. 
It is worth mentioning that VARGAN performance is not dependent on the length of conditions. 
The variation in samples generated by VARGAN is encouraged by another factor (i.e., MCR values), which removes the limitation to control the length and number of categories.

\section{Conclusions}
\label{sec:conclusion}
In this paper, we proposed a new GAN architecture called VARGAN to alleviate the mode collapse issue. 
VARGAN incorporates a new additional network that measures the samples' diversity. The new network's loss on generated samples is used to penalize the generator and to introduce diversity in the generated samples. 
We compared the performance of our method with the state-of-the-art GAN architectures on three different datasets. 
Results show a high performance for the proposed VARGAN architecture, fast convergence and low training times. 
We also investigated the effect of conditioning on mode collapse in GANs. 
Our experiments indicate that conditioning reduces the number of generated modes and induces mode collapse, which can be a cause of deterministic sample generation based on the auxiliary information. 
We also experimented with MSGAN, which is proposed for reducing mode collapse in conditional GANs. 
Our results show that, among the conditional models, for feed-forward structures, conditional VARGAN shows slightly better performance compared to MSGAN, whereas, for convolutional structures, MSGAN outperforms conditional VARGAN. 
Furthermore, we examine the effect of conditioning length on the mode collapse. 
Our analysis with EES dataset shows that mode collapse happens when generator ignores the noise randomness, and creates deterministic results based on conditions. 
We observe a resistance of VARGAN to the condition vector length, which points out to the different architecture of VARGAN compared to the existing models. 


Our study contributes to a better understanding of methodologies to address mode collapse issue, however, evaluating mode collapse remains a challenging task. 
In this study, we limit the evaluation of the experiments to the performance metrics discussed in previous studies on mode collapse in GANs. 
In our analysis, we solely focus on one network architecture extracted from other studies in the literature, which may affect the models' performance. 
In future research, we plan to design more extensive hyperparameter tuning experiments to check the model's capabilities across different architectures. 
We also have selected one architecture for our VarNet model inspired by \citet{lin2018pacgan}, which can be further improved to better serve the purpose of diversity evaluation. 
Specifically, we can integrate a computational loss based on the designs' differences to our generator's loss function.

 \bibliographystyle{spbasic} 
 \bibliography{refs}

\appendix
\section{Selection of MCR formulation and its parameters}\label{ap:Limited_data_formula}
\begin{figure}[!ht]
    \centering
    \includegraphics[width=0.9\textwidth]{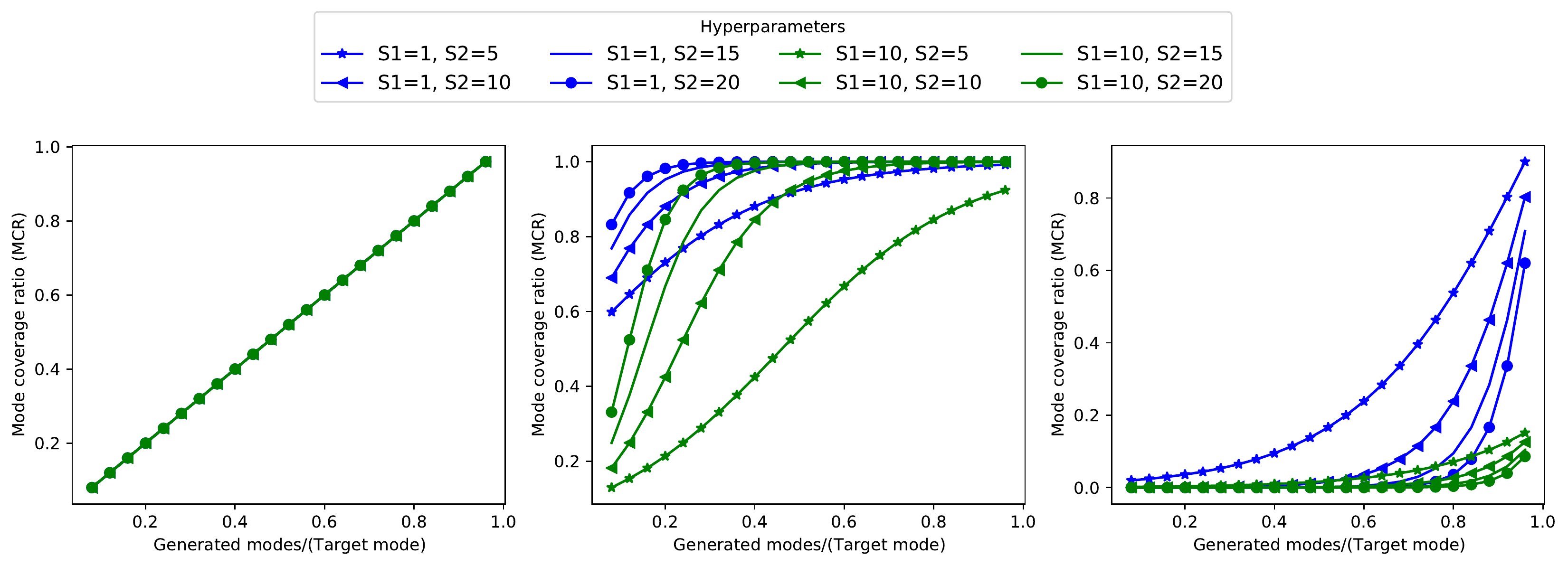}
    \caption{MCR value based on number of generated modes divided by target modes for different equations. Left, middle and right figures in order present Equation~\eqref{eq:mcr-formula_lin}, ~\eqref{eq:mcr-formula_exp} and ~\eqref{eq:mcr-formula_exp_dif}. }
    \label{fig:MCR_fig}
\end{figure}

\begin{figure}[!ht]
\centering
\subfloat[Number of modes \label{fig:36_modes_comp_formulations}]{\includegraphics[width=0.8\textwidth]{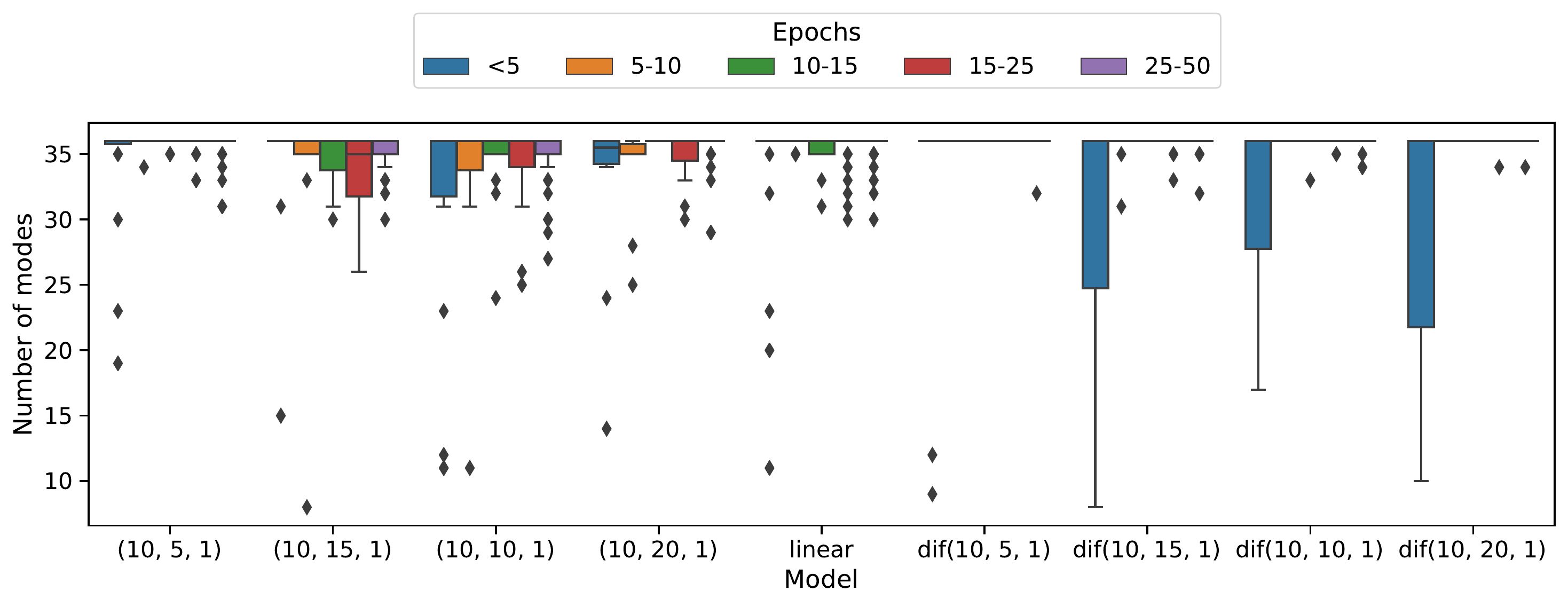}}
\vfill
\subfloat[KL divergence \label{fig:36_comp_KL_formulations}]{\includegraphics[width=0.8\textwidth]{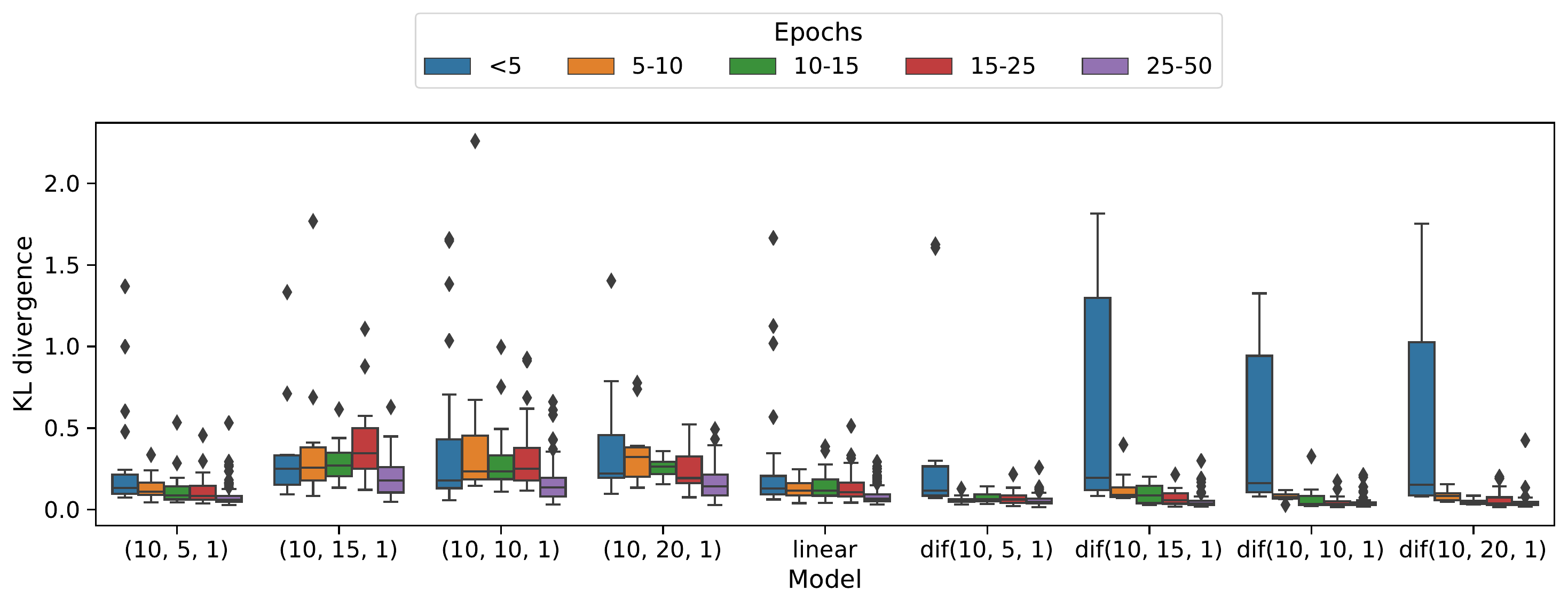}}
\vfill
\subfloat[Percentage of high quality samples \label{fig:36_comp_highq_formulations}]{\includegraphics[width=0.8\textwidth]{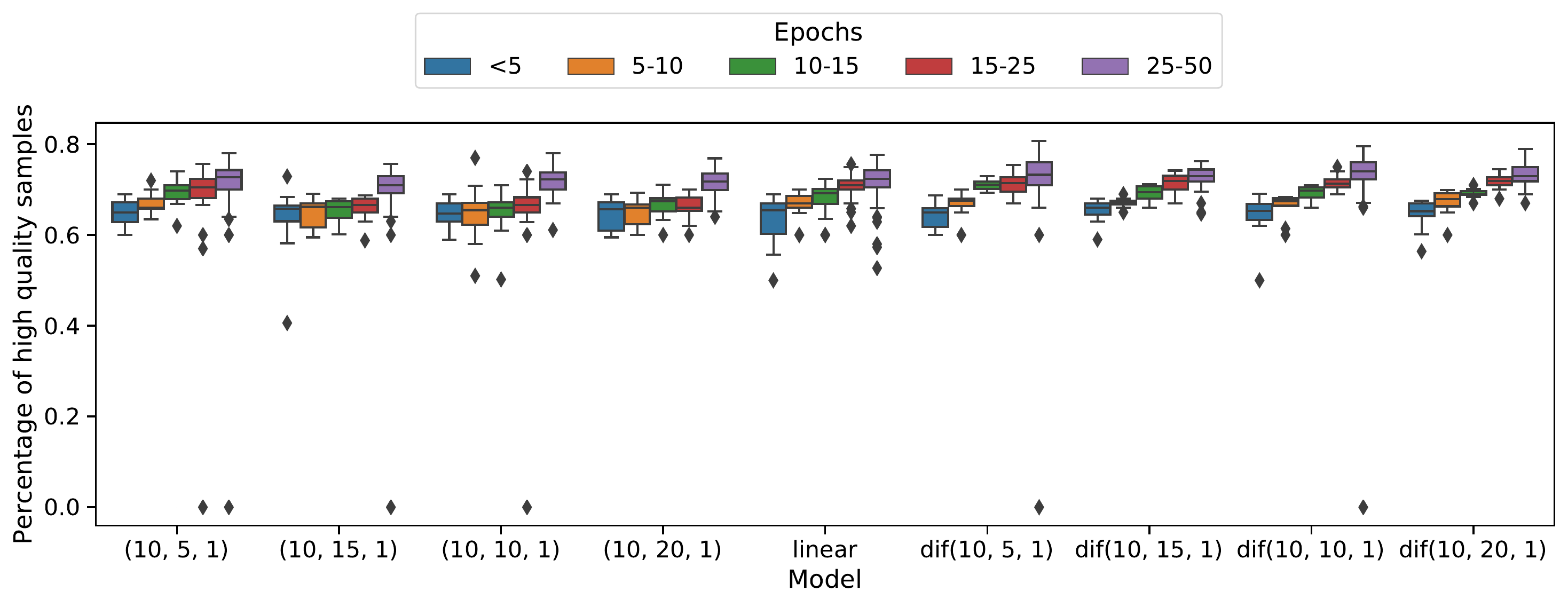}}
\caption{Comparison of different MCR formulations on synthetic 2D grid data with 36 modes averaged over 5 repeats}
\label{fig:Synthetic_result_36grid_boxplot_comp_formulations}
\end{figure}

We have explored three different formulations for the MCR values.
Equation~\eqref{eq:mcr-formula_lin} presents a linear relationship between MCR value and percentage of covered modes.
\begin{equation}
    \label{eq:mcr-formula_lin}
      MCR =  
      \begin{cases}
    \frac{n}{N} & n = 2,\hdots, N-1,  \enspace n \mid B\\ 
      0 & n = 1\\
      
    \end{cases}  
\end{equation}
Equation~\eqref{eq:mcr-formula_exp} models a relationship that achieves early convergence to reward the model with low generator error.
\begin{equation}
    \label{eq:mcr-formula_exp}
      MCR =  
      \begin{cases}
      \frac{\mathrm{L}}{\mathrm{1} +  S_1 \cdot e^{-\frac{n}{N}  \cdot S_2}}  & n = 2,\hdots, N-1,  \enspace n \mid B\\ 
      0 & n = 1\\
      
    \end{cases}  
\end{equation}
Equation~\eqref{eq:mcr-formula_exp_dif} models a relationship where it avoids rewarding the generator too early to keep it motivated.
\begin{equation}
    \label{eq:mcr-formula_exp_dif}
      MCR =  
      \begin{cases}
      \frac{\mathrm{L}}{\mathrm{1} +  S_1 \cdot e^{-\frac{n-N}{N}  \cdot S_2}}  & n = 2,\hdots, N-1,  \enspace n \mid B\\ 
      0 & n = 1\\
    \end{cases}  
\end{equation}
Figure~\ref{fig:MCR_fig} shows the trajectory of different formulations based on the percentage of covered modes.
Firstly, we have experimented with constant values of $L$, $S_1$ and $S_2$ for different formulations on synthetic data with 36 modes. Our results illustrated in Figure~\ref{fig:Synthetic_result_36grid_boxplot_comp_formulations} indicate that Equation~\eqref{eq:mcr-formula_exp} is presenting a suitable trajectory for MCR values over the percentage of covered modes.

\begin{figure}[!ht]
\centering
\subfloat[Number of modes \label{fig:36_modes_comp_exp_formula}]{\includegraphics[width=0.8\textwidth]{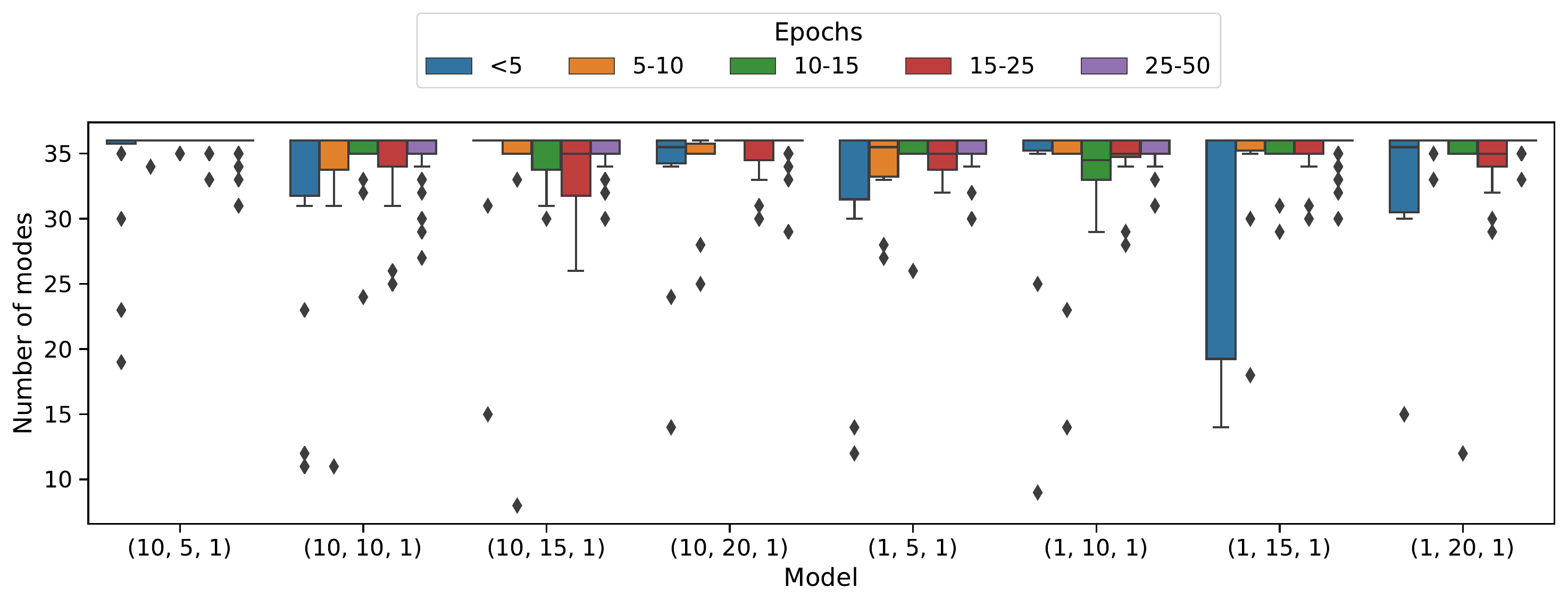}}
\vfill
\subfloat[KL divergence \label{fig:36_comp_KL_exp_formula}]{\includegraphics[width=0.8\textwidth]{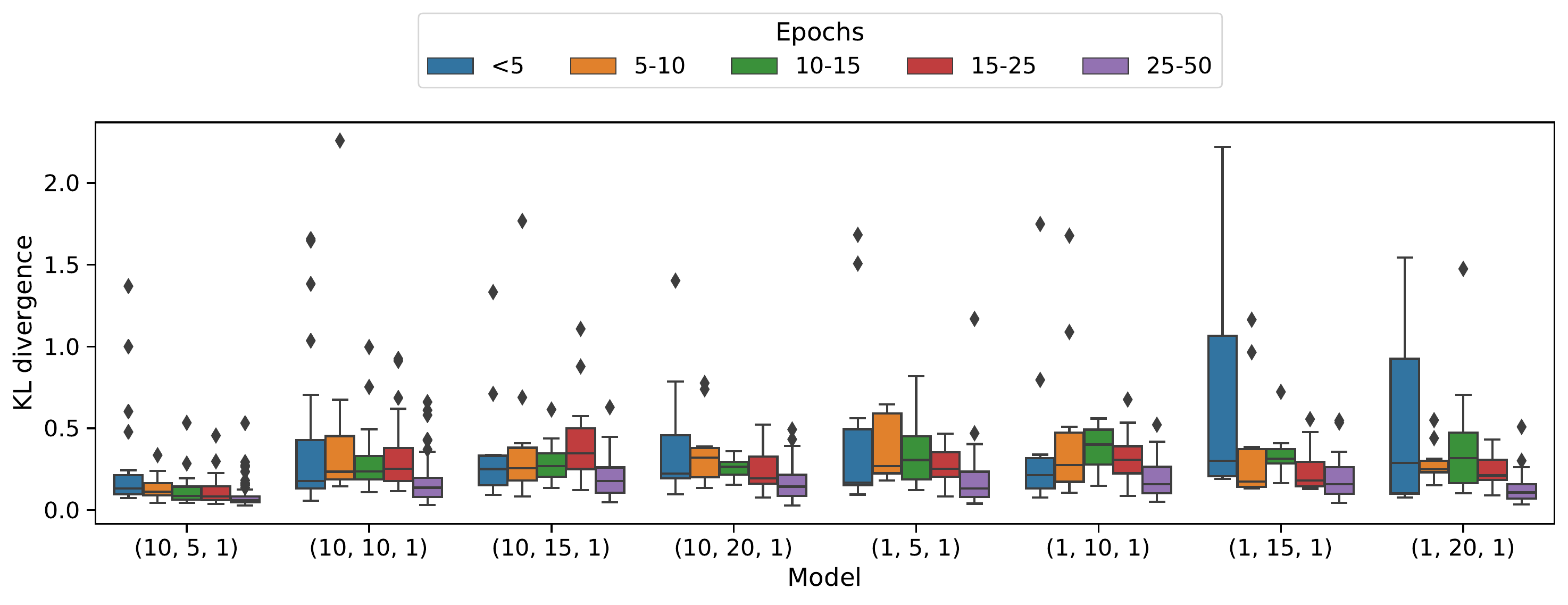}}
\vfill
\subfloat[Percentage of high quality samples \label{fig:36_comp_highq_exp_formula}]{\includegraphics[width=0.8\textwidth]{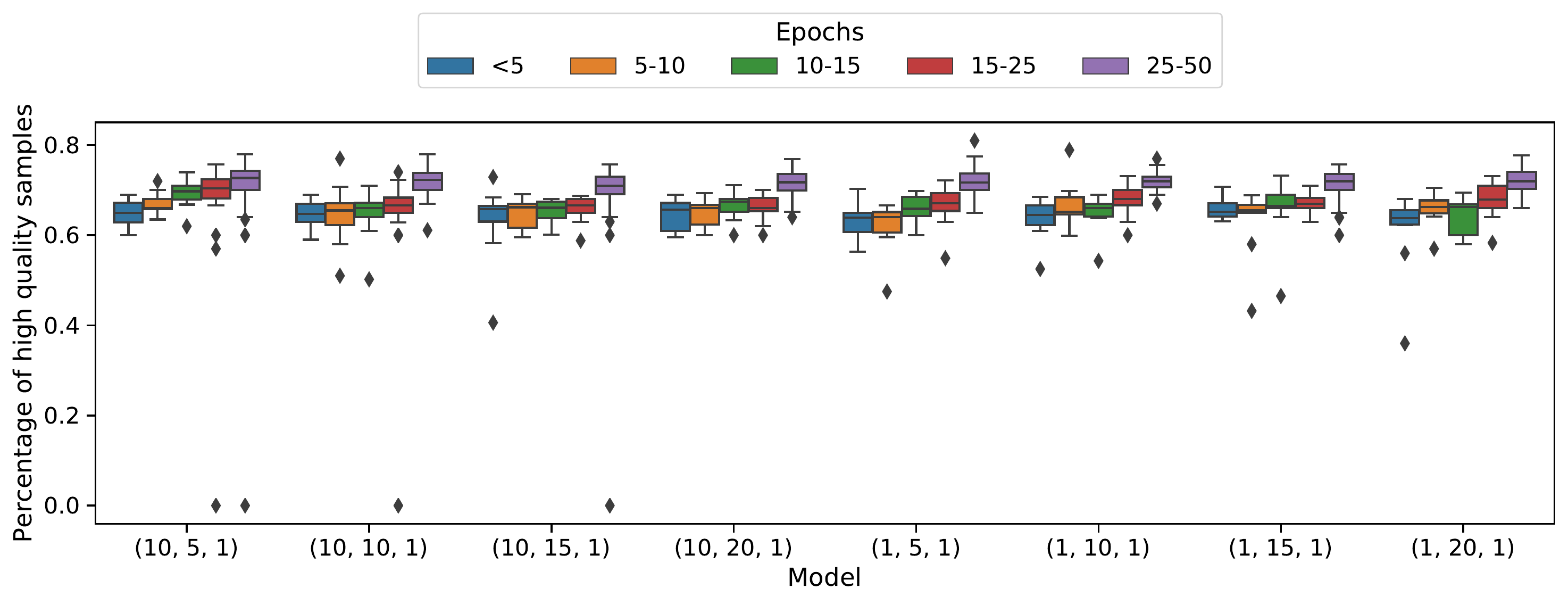}}
\caption{Comparison of different hyperparameters for Equation~\eqref{eq:mcr-formula_exp} on synthetic 2D grid data with 36 modes averaged over 5 repeats}
\label{fig:Synthetic_result_36grid_boxplot_comp}
\end{figure}

Finally, we have used different values of $L$, $S_1$ and $S_2$ on synthetic data with 36 modes. The hyperparameter optimization results are illustrated in Figure~\ref{fig:Synthetic_result_36grid_boxplot_comp}. 
Equation~\eqref{eq:mcr-formula_exp} with $L$, $S_1$ and $S_2$ of 1, 10 and 5 shows an early convergence for all the metrics compared to other formulations and hyperparameters. As shown in Figure~\ref{fig:MCR_fig}, $L$, $S_1$ and $S_2$ of 1, 10 and 5 enforce a low initial MCR value with moderate speed of convergence to MCR value of one. In other words, VARGAN performance does not improve by defining a large MCR value for initial low mode coverage or a fast convergence trajectory.

\section{Architecture of models}\label{ap:model_arch}
In this section, a detailed summary of GAN architectures is presented. Table~\ref{tbl:model_struct_synthetic} presents the feed-forward model architecture used for all the datasets. 
\setlength{\tabcolsep}{3pt}
\renewcommand{\arraystretch}{1.4}
\begin{table}[!ht]
\begin{footnotesize}
\caption{{\footnotesize  Structure of FF GAN model}}
\label{tbl:model_struct_synthetic}
\begin{center}
\scalebox{0.9}{
\noindent\adjustbox{max width=\textwidth}{

\begin{tabular}{ccccccc}
\toprule
Network &
Number of units & Activation function & Regularization\\
\toprule
\multirow{5}{*}{Generator} & 512 & LeakyReLU & -\\
& 1,024 & LeakyReLU & -\\
& 2,048 & LeakyReLU & -\\
& 4,096 & LeakyReLU & -\\
& Output size & Sigmoid & -\\
\toprule
\multirow{5}{*}{Discriminator} & 2,048 & LeakyReLU & -\\
& 1,024 & LeakyReLU & Dropout(0.3)\\
& 512 & LeakyReLU & Dropout(0.3)\\
& 1 & Sigmoid & -\\
\bottomrule
\end{tabular}}

}
\end{center}
\end{footnotesize}
\end{table}
Table~\ref{tbl:model_struct} shows the convolutional model architecture used for stacked MNIST dataset. We have modified the architecture for other GAN variants to implement the specific details of their methodology.
\setlength{\tabcolsep}{3pt}
\renewcommand{\arraystretch}{1.4}
\begin{table}[!ht]
\begin{footnotesize}
\caption{{\footnotesize Structure of convolutional GAN model for stacked MNIST dataset}}
\label{tbl:model_struct}
\begin{center}
\scalebox{0.9}{
\noindent\adjustbox{max width=\textwidth}{

\begin{tabular}{ccccc}
\toprule
 Network &Layer &  \# of channels & Kernel size& Activation function\\
\toprule

\multirow{4}{*}{Generator}& Linear & - & - & ReLU\\
& Conv & 256 & 4 & ReLU\\
&  Conv & 128 & 4& ReLU\\
& Conv & 64 & 4 & ReLU\\
&  Conv & 3 & 4 & Tanh\\
\toprule
\multirow{5}{*}{Discriminator} 
&  Conv & 64 & 4 & LeakyReLU\\
& Conv & 128& 4 & LeakyReLU\\
& Conv& 256 & 4 & LeakyReLU \\
& Conv & 512& 4 & LeakyReLU\\
& Linear & - & - &  Sigmoid\\
\toprule
\end{tabular}}





}
\end{center}
\end{footnotesize}
\end{table}
Convolutional model architecture used for EES dataset is reported in Table~\ref{tbl:model_struct_EES}. Number of convolutional layers is changed to implement both $9\times9$ and $19\times19$ designs.
\setlength{\tabcolsep}{3pt}
\renewcommand{\arraystretch}{1.4}
\begin{table}[!ht]
\begin{footnotesize}
\caption{{\footnotesize Structure of convolutional GAN model for EES dataset}}
\label{tbl:model_struct_EES}
\begin{center}
\scalebox{0.9}{
\noindent\adjustbox{max width=\textwidth}{

\begin{tabular}{ccccccc}
\toprule
 Network &Layer &  \# of channels &Kernel size& Activation function\\
\toprule
\multirow{5}{*}{Generator} & Linear & -  & - & LeakyReLU\\
& Conv & 256 & 3 & LeakyReLU\\
&  Con & 128 & 3 & LeakyReLU\\
&  Conv & 1 & 3 & Sigmoid\\

\toprule
\multirow{5}{*}{Discriminator}  & Conv & 64 & 3 & LeakyReLU\\
&  Conv & 128& 3 & LeakyReLU\\
&  Conv & 256 & 3 & LeakyReLU\\
& Linear & -  &- & Sigmoid \\
\bottomrule

\end{tabular}}

}
\end{center}
\end{footnotesize}
\end{table}

\section{Detailed results}\label{ap:detailed_results}
\begin{figure}[!ht]
\centering
\subfloat[Number of modes \label{fig:8_modes}]{\includegraphics[width=0.8\textwidth]{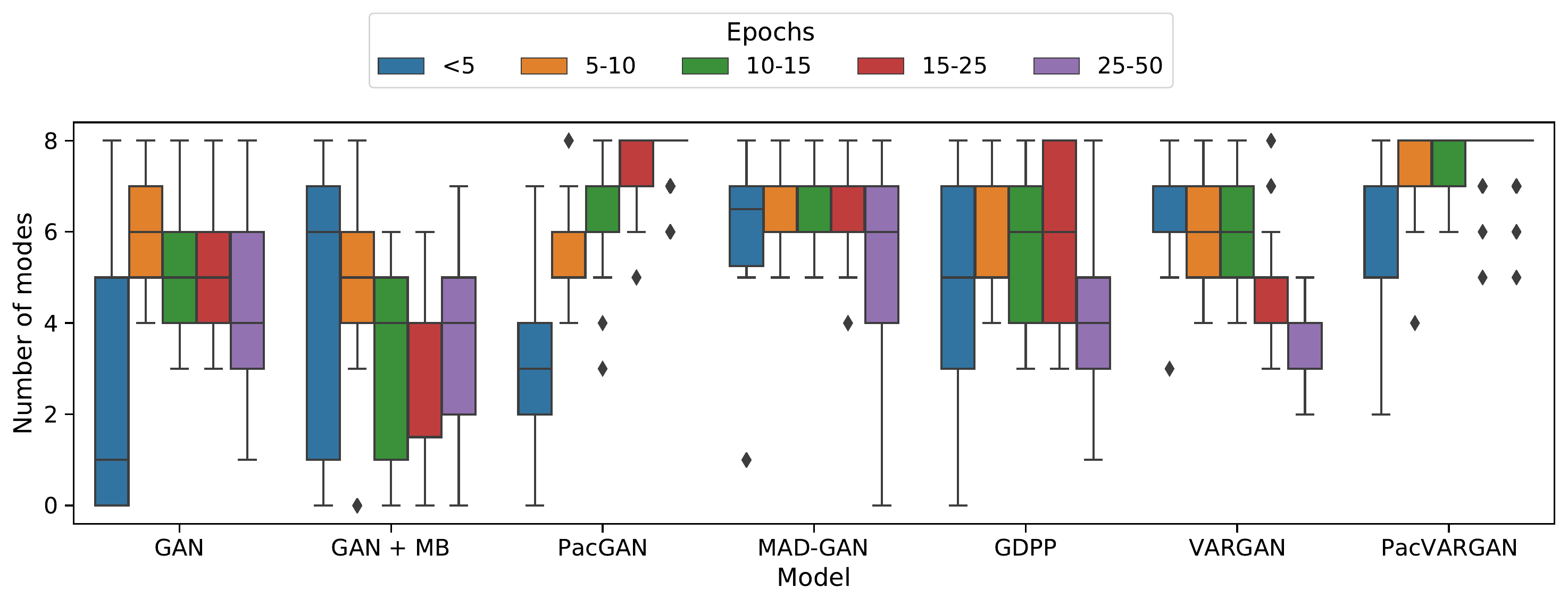}}
\vfill
\subfloat[KL divergence \label{fig:8_KL}]{\includegraphics[width=0.8\textwidth]{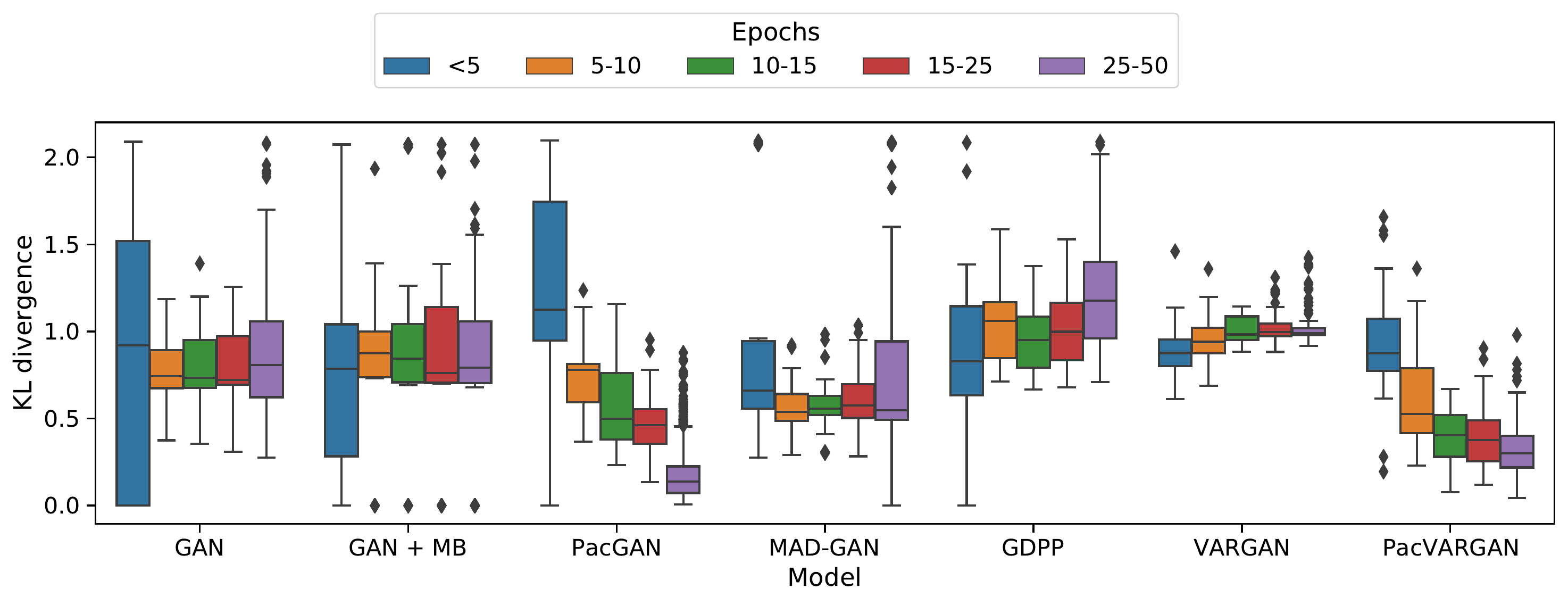}}
\vfill
\subfloat[Percentage of high quality samples \label{fig:8_highq}]{\includegraphics[width=0.8\textwidth]{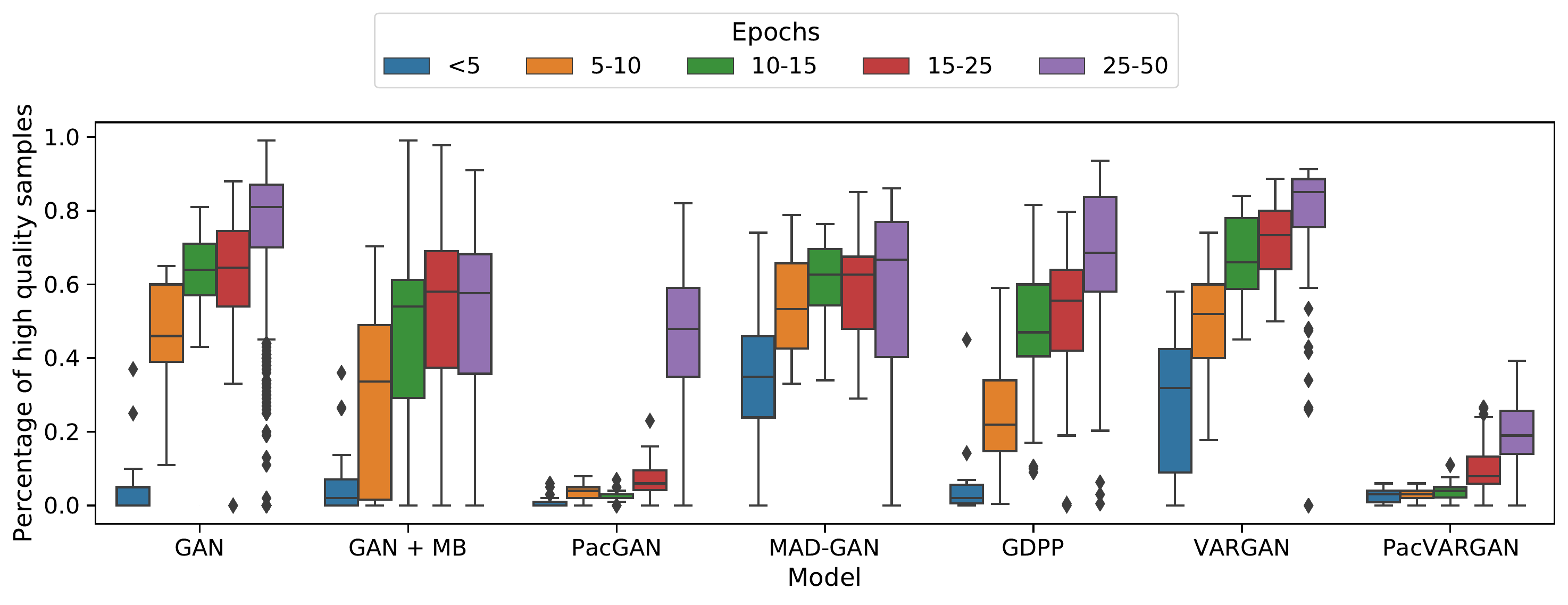}}
\caption{Comparison of different GAN models on synthetic 2D ring data with 8 modes averaged over 5 repeats}
\label{fig:Synthetic_result_8grid_boxplot}
\end{figure}

\begin{figure}[!ht]
\centering
\subfloat[Number of modes \label{fig:25_modes}]{\includegraphics[width=0.8\textwidth]{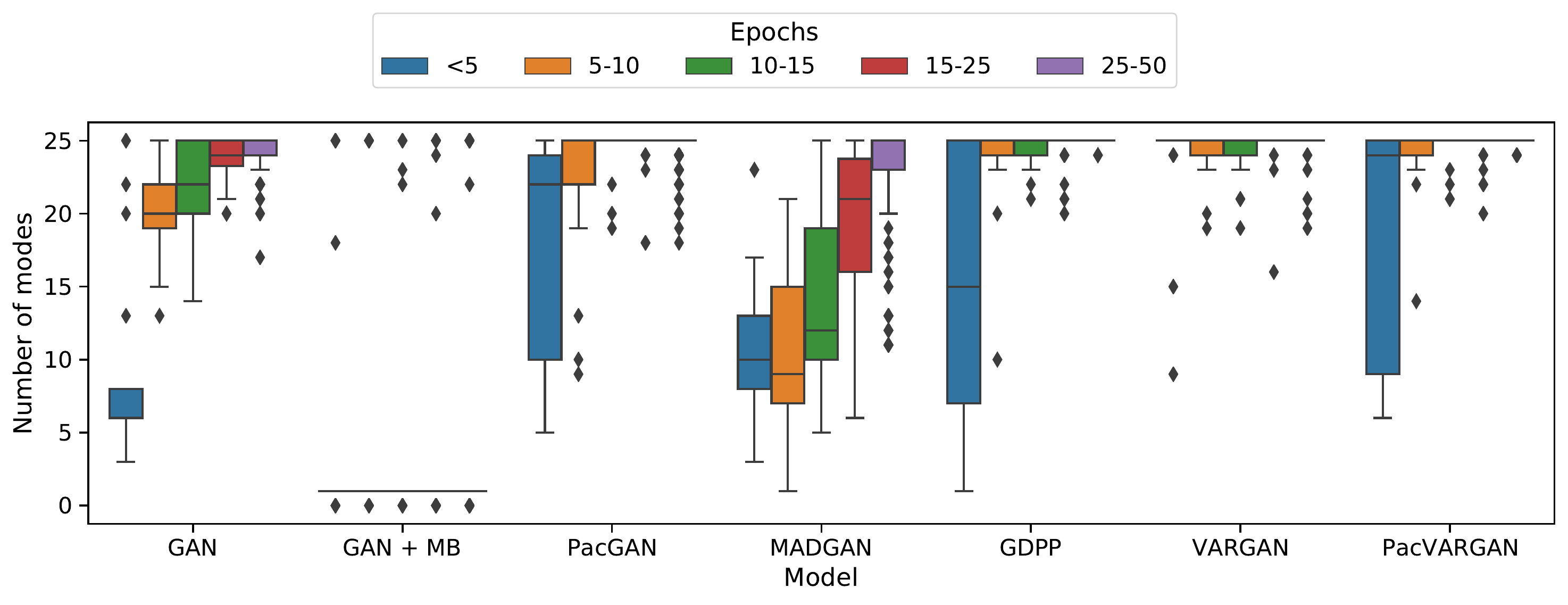}}
\vfill
\subfloat[KL divergence \label{fig:25_KL}]{\includegraphics[width=0.8\textwidth]{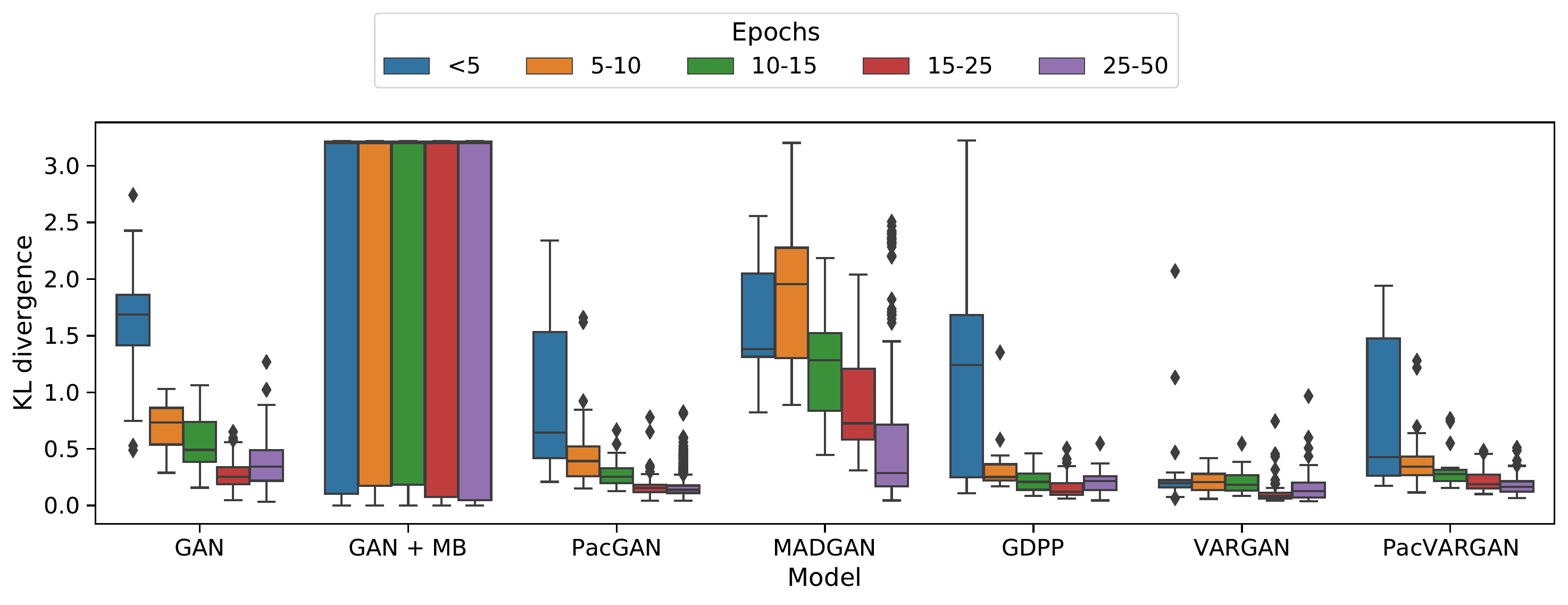}}
\vfill
\subfloat[Percentage of high quality samples \label{fig:25_highq}]{\includegraphics[width=0.8\textwidth]{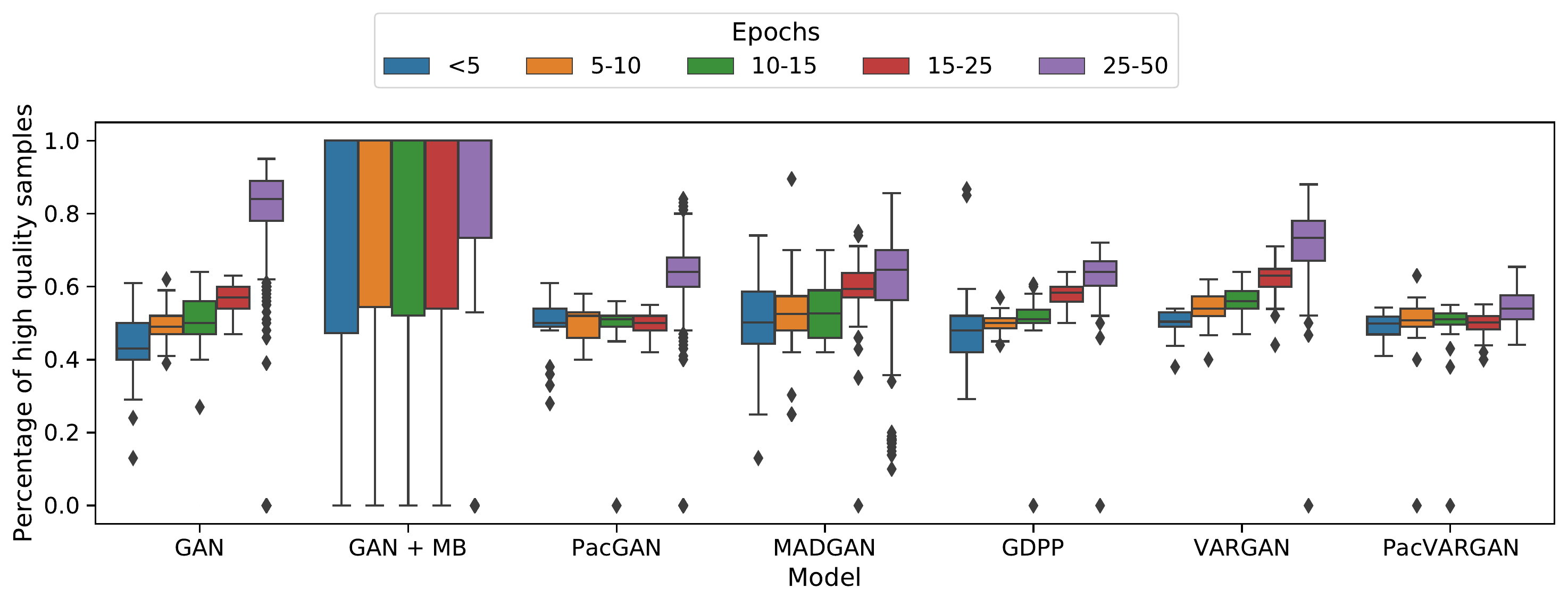}}
\caption{Comparison of different GAN models on synthetic 2D grid data with 25 modes averaged over 5 repeats}
\label{fig:Synthetic_result_25grid_boxplot}
\end{figure}

In this section, performance comparison of GAN models over the epochs is presented. Figure~\ref{fig:Synthetic_result_8grid_boxplot} and ~\ref{fig:Synthetic_result_25grid_boxplot} illustrate the convergence of the models for synthetic data with 8 and 25 modes based on different performance metrics and over epochs. VARGAN and GDPP models seem to have early convergence in the beginning epochs for synthetic data with 8 modes. VARGAN shows great early convergence on synthetic data with 25 modes as well, and the rest of the models follow it by a large gap. 

\begin{figure}[!ht]
\centering
\subfloat[Number of modes \label{fig:comp_modes}]{\includegraphics[width=0.8\textwidth]{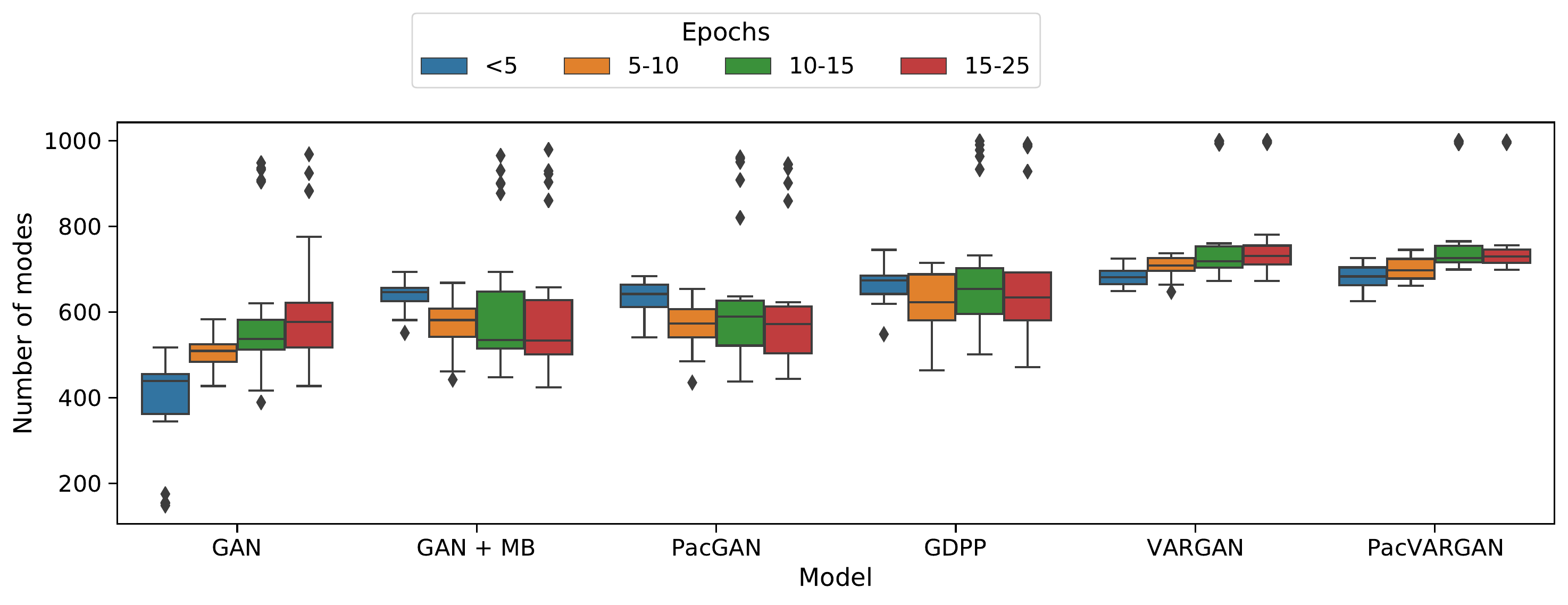}}
\vfill
\subfloat[KL divergence \label{fig:comp_KL}]{\includegraphics[width=0.8\textwidth]{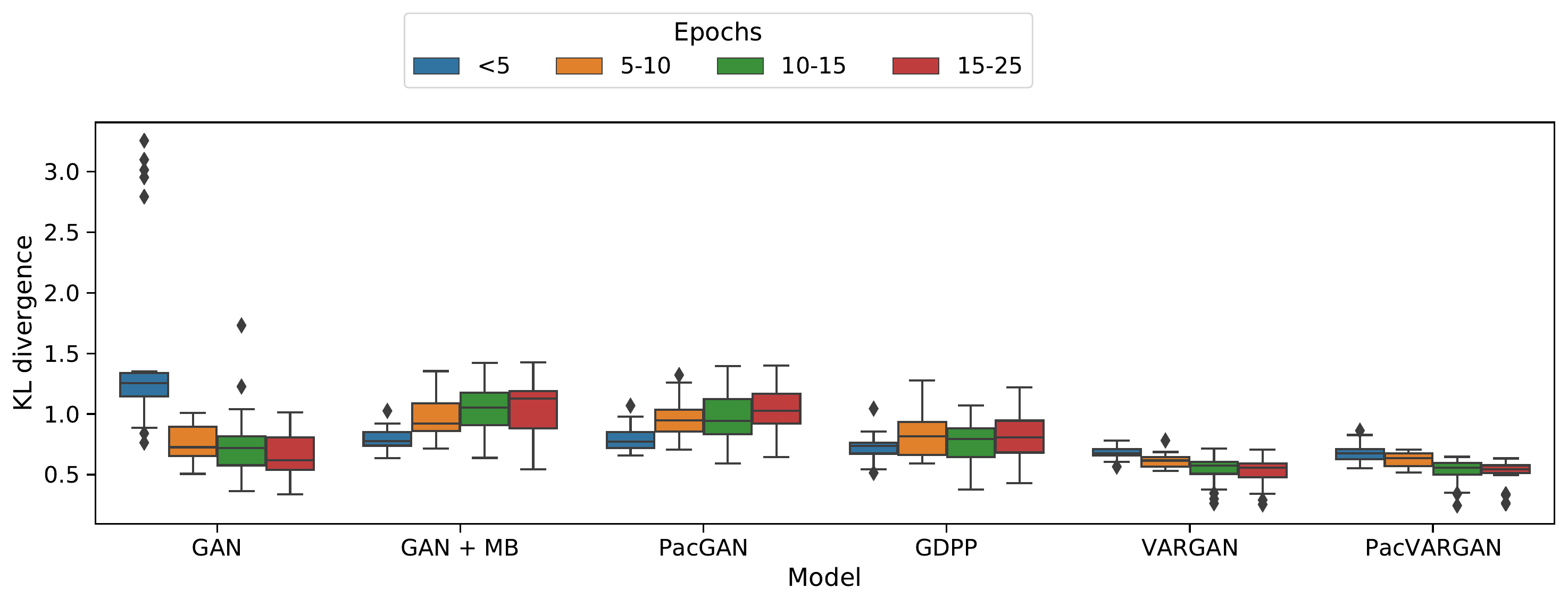}}
\vfill
\subfloat[Inception Score \label{fig:comp_highq}]{\includegraphics[width=0.8\textwidth]{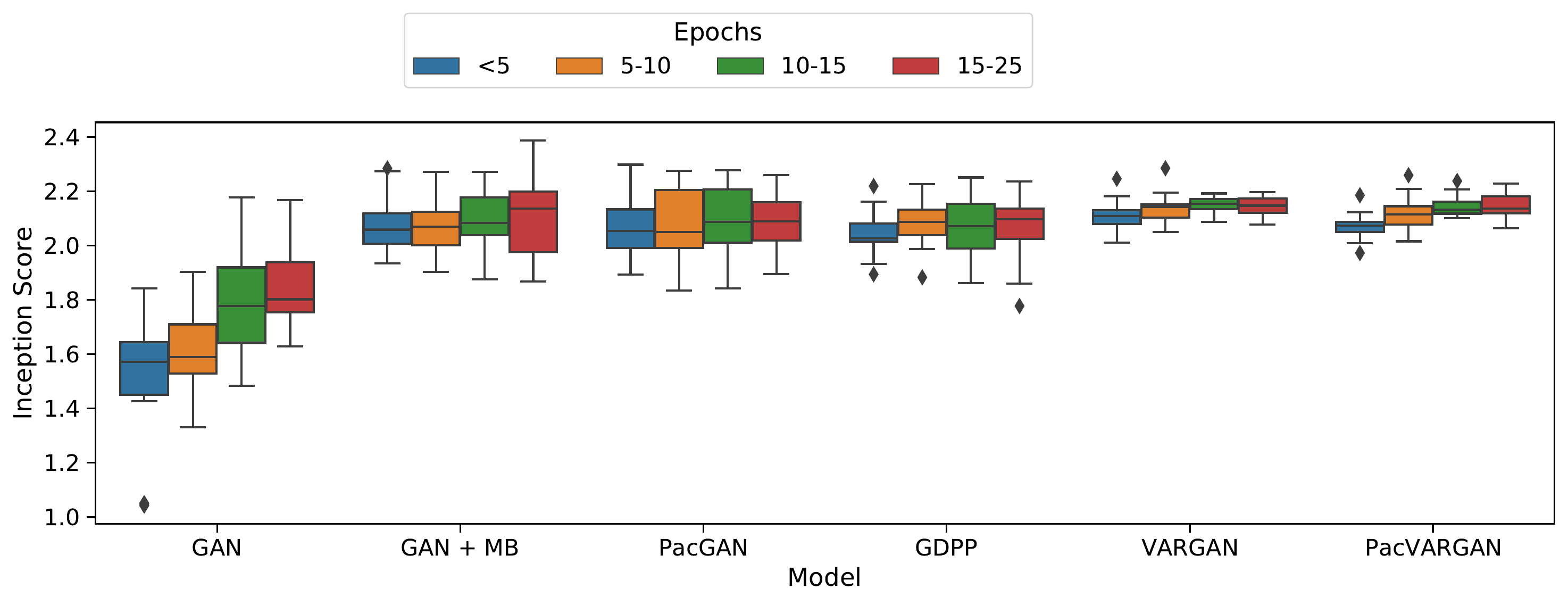}}
\caption{Performance comparison of different GAN models with convolutional GAN structure for stacked MNIST dataset averaged over 5 repeats}
\label{fig:Stacked_MNIST_CNN_boxplot}
\end{figure}

Figure~\ref{fig:Stacked_MNIST_CNN_boxplot} presents the convergence of performance metrics for convolutional GAN models on stacked MNIST data. Both VARGAN and PacVARGAN show great early convergence on all metrics.

\end{document}